\documentclass{article}

\PassOptionsToPackage{numbers,sort,compress}{natbib}

\usepackage[preprint]{Style}

\usepackage[utf8]{inputenc} 
\usepackage[T1]{fontenc}    
\usepackage{textcomp}      
\usepackage{hyperref}       
\usepackage{url}            
\usepackage{booktabs}       
\usepackage{amsfonts}       
\usepackage{nicefrac}       
\usepackage{microtype}      
\usepackage{xcolor}         
\usepackage{xparse}         
\usepackage{graphicx}
\usepackage{amsmath}
\makeatletter
\@ifundefined{And}{\newcommand{\And}{\and}}{\renewcommand{\And}{\and}}
\@ifundefined{AND}{\newcommand{\AND}{\and}}{\renewcommand{\AND}{\and}}
\makeatother
\usepackage{enumerate}
\usepackage{enumitem}
\usepackage{placeins}
\usepackage{multirow}
\usepackage{comment}
\usepackage{threeparttable} 
\usepackage{tabularx}      
\usepackage{makecell}     
\usepackage{bbm} 
\usepackage{bm}
\usepackage{siunitx}

\usepackage{amssymb}
\usepackage{newunicodechar}
\usepackage{etoolbox}

\newunicodechar{μ}{\ensuremath{\mu}}
\newunicodechar{α}{\ensuremath{\alpha}}
\newunicodechar{β}{\ensuremath{\beta}}
\newunicodechar{τ}{\ensuremath{\tau}}
\newunicodechar{λ}{\ensuremath{\lambda}}
\newunicodechar{Φ}{\ensuremath{\Phi}}
\newunicodechar{∆}{\ensuremath{\Delta}}
\newunicodechar{Δ}{\ensuremath{\Delta}}
\newunicodechar{≤}{\ensuremath{\leq}}
\newunicodechar{≥}{\ensuremath{\geq}}
\newunicodechar{≈}{\ensuremath{\approx}}
\newunicodechar{∼}{\ensuremath{\sim}}
\newunicodechar{×}{\ensuremath{\times}}
\newunicodechar{±}{\ensuremath{\pm}}
\newunicodechar{→}{\ensuremath{\to}}
\newunicodechar{−}{\ensuremath{-}}
\newunicodechar{′}{\ensuremath{'}}
\newunicodechar{∈}{\ensuremath{\in}}
\newunicodechar{∑}{\ensuremath{\sum}}
\newunicodechar{⊤}{\ensuremath{\top}}
\newunicodechar{–}{--}
\newunicodechar{—}{---}
\newunicodechar{’}{'}
\newunicodechar{‘}{`}
\newunicodechar{“}{``}
\newunicodechar{”}{''}
\newunicodechar{∗}{\ensuremath{*}}

\definecolor{blackrn}{rgb}{0,0,0}

\RenewDocumentCommand{\textcolor}{o m +m}{%
  \begingroup
  \IfNoValueTF{#1}{\color{#2}}{\color[#1]{#2}}%
  #3%
  \endgroup
}

\newcommand{\safeincludegraphics}[2][]{%
  \IfFileExists{#2}{%
    \includegraphics[#1]{#2}%
  }{%
    \fbox{\begin{minipage}[c][0.28\textheight][c]{0.9\linewidth}\centering
    Missing figure file: \texttt{#2}
    \end{minipage}}%
  }%
}

\title{DIVER-1: Scaling Intracranial EEG Foundation Models for Transferable Representations}

\author{%
  Danny Dongyeop Han\textsuperscript{*,\textdagger,1} \\
  Seoul National University \\
  Seoul, Republic of Korea \\
  \texttt{dyhan316@snu.ac.kr}
  \And
  Yonghyeon Gwon\textsuperscript{*,1} \\
  Seoul National University \\
  Seoul, Republic of Korea
  \And
  Ahhyun Lucy Lee\textsuperscript{*,1} \\
  Seoul National University \\
  Seoul, Republic of Korea
  \And
  Taeyang Lee\textsuperscript{*,1} \\
  Seoul National University \\
  Seoul, Republic of Korea
  \AND
  Seong Jin Lee\textsuperscript{1} \\
  Seoul National University \\
  Seoul, Republic of Korea
  \And
  Jubin Choi\textsuperscript{1} \\
  Seoul National University \\
  Seoul, Republic of Korea
  \And
  Sebin Lee\textsuperscript{1} \\
  Seoul National University \\
  Seoul, Republic of Korea
  \And
  Jihyun Bang\textsuperscript{1} \\
  Seoul National University \\
  Seoul, Republic of Korea
  \AND
  Seungju Lee\textsuperscript{1} \\
  Seoul National University \\
  Seoul, Republic of Korea
  \And
  David Keetae Park\textsuperscript{2} \\
  Brookhaven National Laboratory \\
  New York, United States
  \And
  Shinjae Yoo\textsuperscript{2} \\
  Brookhaven National Laboratory \\
  New York, United States
  \AND
  Chun Kee Chung\textsuperscript{\textdagger,1} \\
  Seoul National University \\
  Seoul, Republic of Korea \\
  \texttt{chungc@snu.ac.kr}
  \And
  Jiook Cha\textsuperscript{\textdagger,1} \\
  Seoul National University \\
  Seoul, Republic of Korea \\
  \texttt{connectome@snu.ac.kr}
}

\begin{document}

\maketitle

\begin{center}
\textsuperscript{*} Equal contribution \qquad
\textsuperscript{\textdagger} Corresponding authors
\end{center}

\begin{abstract}

Intracranial EEG (iEEG) provides direct, millisecond-scale recordings of human neural activity, but reusable representation learning is difficult because electrode layouts, anatomical coverage, referencing schemes, and recording conditions vary across patients and centers. We introduce DIVER-1, a self-supervised iEEG foundation model for variable-input recordings that combines any-variate electrode--time attention, spatio-temporal resampling, input-conditioned positional embeddings, and multi-domain masked reconstruction without assuming a fixed electrode montage. We pretrain two variants, DIVER-1-0.1s and DIVER-1-1s, on 5,310 hours of ECoG and SEEG spanning 352k channel-hours, roughly 54$\times$ the BrainTreeBank-based pretraining volume. We evaluate DIVER-1 on two held-out benchmarks: Neuroprobe for naturalistic cognitive decoding and MAYO for seizure detection. On leakage-aware Neuroprobe, DIVER-1-0.1s outperforms prior evaluated iEEG foundation models despite using no BrainTreeBank recordings---the corpus underlying Neuroprobe---during pretraining; it also exceeds the linear spectrogram decoder in mean AUROC and remains competitive with stronger nonlinear baselines, a level prior evaluated iEEG foundation models did not reach. DIVER-1-1s also achieves the top AUROC on MAYO seizure detection. Finally, we conduct, to our knowledge, the first controlled compute-aware scaling study for self-supervised iEEG pretraining, sweeping data scale, subject count, training duration, and model size up to 1.8B parameters. Our results indicate a data-constrained regime: expanding unique recordings and training sufficiently long are more reliable scaling axes than increasing parameter count alone. Code is available at \href{https://anonymous.4open.science/r/DIVER-1_update}{link}.

\end{abstract}

\section{Introduction}

Large-scale self-supervised pretraining has enabled foundation models across language, vision, and speech: models trained on broad unlabeled corpora to learn reusable representations for many downstream tasks~\textcolor{black}{\citep{bommasani2021opportunities}}. In neuroscience, the analogous goal is an encoder that learns from long unlabeled neural recordings and transfers across subjects, tasks, and clinical centers. Intracranial EEG (iEEG) is a compelling testbed for this goal: electrodes placed on or within the brain record field potentials close to their sources, preserving high-frequency, millisecond-scale neural dynamics that are attenuated and spatially blurred in non-invasive EEG~\citep{buzsaki2012origin,zahorodnii2025neuroprobe,oganesian2025barista}. These properties make iEEG valuable for cognitive decoding, brain--computer interfaces~\citep{card_speech_2024,wairagkar_voice_2025}, and precision neuromedicine~\textcolor{black}{\citep{scangos2021closed, khambhati2019functional}}.

Yet current iEEG foundation models have not fully realized this promise: under leakage-aware evaluation, simple spectrogram-based linear decoders often outperform pretrained iEEG models~\citep{zahorodnii2025neuroprobe}. The core difficulty is that iEEG is clinically implanted and patient-specific, so recordings vary in channel count, anatomical coverage, electrode type, reference scheme, and recording center. Thus, general iEEG foundation modeling requires more than standard Transformer pretraining; it requires addressing three bottlenecks in architecture, input interface, and scale.

\textbf{Bottleneck 1: restricted attention architectures for spatio-temporal neural dynamics.} Transformer-based foundation models model interactions through attention between patch tokens, i.e., local segments of the spatio-temporal neural recording. However, self-attention is order-agnostic by default: without additional structure, it does not know whether two tokens are adjacent in time or from the same electrode. This is important for iEEG, where behaviorally relevant activity can involve both local temporal dynamics within an electrode and coordination across distributed electrodes.

Existing iEEG models address this with different trade-offs. BrainBERT~\citep{wang2023brainbert} models electrodes independently, precluding modeling direct cross-channel interaction. Brant~\citep{zhang2023brant} and PopT~\citep{chau2025popt} introduce spatial modeling but separate temporal and spatial processing. These designs are efficient, but cross-channel-time interactions remain indirect: activity at electrode $a$ and time $t$ can influence activity at electrode $b$ and time $t+\Delta$ only through staged or restricted attention routes. This is potentially limiting for modeling the brain's delayed inter-regional coordination, phase-lagged coupling, and distributed oscillatory communication~\citep{deco2011emerging,stam2007phase,fries2005mechanism,varela2001brainweb}.

Full spatio-temporal attention is therefore attractive because every electrode--time token can attend to every other token, but it still needs structure identifying each token's temporal and channel origin. BaRISTA~\citep{oganesian2025barista} uses full spatio-temporal attention and injects temporal order directly into attention with RoPE, but adds spatial information as input-level metadata embeddings. This leaves same- versus cross-channel relations implicit in token features, despite \citet{su2024roformer} suggesting that relational structure can be more effective when encoded in attention rather than added to inputs~\citep{su2024roformer}. Moreover, many iEEG datasets also lack complete channel metadata~\citep{bbrinkm2014upenn,carzaniga2025swec}, so a general encoder should remain channel-relation aware even without spatial metadata embeddings.

\textbf{Bottleneck 2: fixed pretraining interfaces for variable iEEG inputs.} iEEG foundation models must also handle heterogeneity in the input interface itself. Clinical iEEG varies across patients in channel count, anatomical coverage, and electrode sub-modality (SEEG/ECoG). Downstream tasks also require different temporal contexts, from sub-second decoding of rapid perceptual or linguistic events to multi-second clinical windows in seizure detection. A general iEEG backbone should therefore not assume a single native channel count, electrode sub-modality, or task timescale.

Existing iEEG foundation models are trained through narrower interfaces. BrainBERT, Brant, PopT, and BaRISTA~\citep{wang2023brainbert,zhang2023brant,chau2025popt,oganesian2025barista} are pretrained on SEEG and use fixed pretraining input windows, creating a mismatch with ECoG recordings and with downstream tasks that spans shorter or longer windows. Furthermore, channel semantics depend not only on anatomical location, but also on the rest of the implant and the reference scheme. Under common-average referencing, for example, each contact is measured relative to the average of that patient's available contacts. Thus, a temporal-lobe contact in a temporal-focused implant is referenced against a different signal mixture than a comparable contact in a broader implant spanning temporal and extra-temporal regions. In this sense, a channel's meaning depends on the implant as a set, not only on its coordinate. EEG methods such as ACPE~\citep{wang2024cbramod} improve input adaptivity through conditional positional encodings, but channel-axis convolutions make the encoding order-sensitive: permuting the same electrodes changes the convolutional neighborhood even though the recording set is unchanged. For iEEG, where channel order is arbitrary across patients and centers, the input interface should preserve channel permutation equivariance.

\textbf{Bottleneck 3: under-scaled pretraining and missing iEEG scaling laws.} Current iEEG pretraining remains narrow and small. BrainBERT, PopT, and BaRISTA ~\citep{wang2023brainbert,chau2025popt,oganesian2025barista} train on BrainTreeBank~\citep{wang2024brain}, and are evaluated on downstream tasks from the same dataset~\citep{zahorodnii2025neuroprobe}. This enables controlled benchmarking, but limits pretraining to pediatric movie-watching recordings at modest scale, roughly 40--43 hours total. It also differs from the intended use of foundation-models, where transfer should span subjects, centers, age groups, behavioral states, and recording protocols.Clinical monitoring is uniquely valuable because implanted electrodes can remain in place for days, yielding long recordings during naturalistic hospital behavior~\citep{bbrinkm2014upenn}. Brant~\citep{zhang2023brant} is an important scale exception, but its fixed 6 s patch interface limits use in sub-second contexts such as rapid perceptual and linguistic tasks.

Once pretraining data expand, compute allocation becomes the next bottleneck. Under a fixed compute budget, should iEEG foundation models scale data, parameters, training duration, or subject diversity? In language and vision, scaling-law studies have made such decisions less heuristic~\citep{kaplan2020scaling,hoffmann_chinchilla_2022,data_constrained_scaling_muennighoff2023scaling}. For neural data, recent scaling studies have focused mainly on supervised decoding or participant diversity in specific task settings~\citep{banville2025scalinglaws,bomatter2025limitedparticipantdiversity}, while EEG/iEEG foundation models provide only coarse evidence that larger datasets or models improve performance~\citep{zhang2023brant,wang2024cbramod,jiang2024labram}. The field therefore still lacks controlled, compute-aware sweeps over data, model size, training duration, and subject diversity for self-supervised iEEG foundation-model pretraining.

\textbf{DIVER-1: an iEEG-specific architecture and scaling study.} We address these three bottlenecks with DIVER-1, a self-supervised iEEG foundation model that combines any-variate electrode--time attention (Bottleneck 1), input-conditioned encoding and spatio-temporal resampling for variable layouts (Bottleneck 2), and large-scale heterogeneous pretraining paired with a controlled scaling-law study (Bottleneck 3). Our contributions are:

\begin{itemize}[itemsep=0pt, topsep=0pt, parsep=0pt]
    \item We introduce \textbf{DIVER-1}, a self-supervised iEEG foundation model for variable electrode--time inputs. Its any-variate attention encodes temporal order and same-/cross-channel structure within attention itself, enabling direct cross-channel, cross-time modeling. STR (Spatio-Temporal Resampling) and STCPE (Spatio-Temporal Conditional Positional Embedding) make the input interface robust to variable electrode layouts and temporal contexts, while MDRO trains the model to reconstruct masked patches across complementary signal domains. We instantiate two patch-scale variants: DIVER-1-0.1s and DIVER-1-1s.

    \item We demonstrate \textbf{state-of-the-art cross-dataset iEEG decoding from large-scale pretraining}.
By pretraining on 5,310 hours of iEEG---over $54\times$ more channel-hours than BrainTreeBank-based pretraining---DIVER-1 transfers across 16 downstream evaluations spanning disjoint naturalistic (Neuroprobe) and clinical (MAYO) iEEG benchmarks. On BrainTreeBank-derived Neuroprobe, DIVER-1 achieves the best mean AUROC despite no BrainTreeBank pretraining, reversing the prior finding that iEEG foundation models pretrained on BrainTreeBank itself underperform linear spectrogram baselines.
    
    \item We provide, to our knowledge, the \textbf{first controlled compute-aware scaling-law study} for self-supervised iEEG foundation-model pretraining. Sweeping data scale, model size up to 1.82B parameters, training duration, compute, and subject count, we find that iEEG pretraining follows a data-constrained scaling regime, yielding a practical recipe for future iEEG foundation models: scale data first, train longer second, and scale parameters last.

\end{itemize}

\section{Methods}

\subsection{Model architecture}

\begin{figure}[t]
    \centering
    \includegraphics[width=0.9\linewidth]{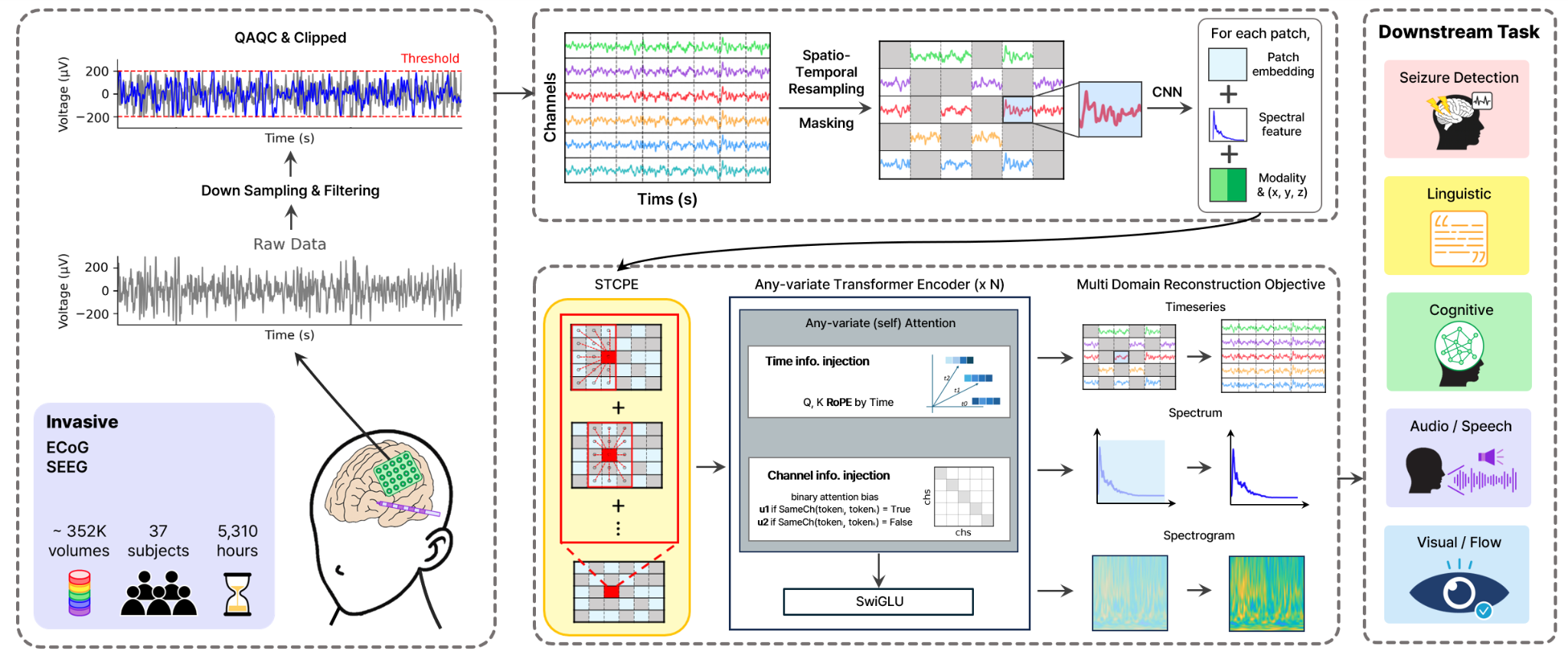}
    \caption{\textbf{DIVER-1 architecture, pretraining, and downstream evaluation pipeline.} After preprocessing, input patches are randomly masked and enhanced by adding modality, spectral, and CNN patch encodings, along with STCPE. The patches are then processed through any-variate Transformer encoders and trained to reconstruct missing patches across multiple signal domains. The pretrained model is then applied to diverse downstream tasks.} 
    \label{fig:Arch}
\end{figure}

As illustrated in Figure~\ref{fig:Arch}, DIVER-1 patchifies each iEEG channel, encodes patches with a temporal CNN, and forms the token grid \(\mathbf{X} = \mathbf{Y}_{\mathrm{CNN}} + \mathbf{E}_{\mathrm{STCPE}} + [\mathbf{E}_{\mathrm{position}}, \mathbf{E}_{\mathrm{modality}}] + \mathbf{E}_{\mathrm{spectral}}\) where \([\cdot,\cdot]\) denotes concatenation and the embedding terms are defined below. 
A MOIRAI-based any-variate Transformer processes \(\mathbf{X}\) across variable iEEG electrode layouts. 
We train two patch-size variants, DIVER-1-1s and DIVER-1-0.1s. 
Architectural hyperparameters are provided in Table~\ref{tab:diver_hyperparams}.

\paragraph{Patch encoding ($\mathbf{Y}_{\text{CNN}}$).}

Given an input $\mathbf{Y}\in\mathbb{R}^{C\times T}$, we split each channel into non-overlapping temporal patches of length $P$, where $P=500$ for DIVER-1-1s and $P=50$ for DIVER-1-0.1s at 500\,Hz. A three-layer CNN then extracts local features from each patch, yielding $\mathbf{Y}_{\mathrm{CNN}}\in\mathbb{R}^{C\times N\times d_{\mathrm{model}}}$, where $N=T/P$ is the number of patches and $d_{\text{model}}$ is the embedding dimension.

\paragraph{Spatio-Temporal Conditional Positional Embedding (STCPE) ($\mathbf{E}_{\text{STCPE}}$).}

STCPE injects \textbf{input-conditioned local spatio-temporal positional bias} without assuming a
fixed iEEG montage. Because iEEG voltages are reference-dependent, a channel's signal reflects both its 3D location and its coupling to other channels through the reference scheme. ACPE~\citep{wang2024cbramod} captures such local
channel--time context with convolutions, but its channel-axis convolution makes the encoding
channel order-sensitive. STCPE replaces this with a sliding-window any-variate Transformer block (MOIRAI), yielding
\textbf{temporally translation-equivariant} and \textbf{channel-permutation-equivariant} positional bias.

Let $\tilde{\mathbf{X}} = P_{\downarrow}(\mathbf{X})$, where
$P_{\downarrow}: \mathbb{R}^{d_{\mathrm{model}}}\rightarrow \mathbb{R}^{d_{\mathrm{model}}/8}$, and let
$m=(w-1)/2$ for a temporal window of width $w$. For each window center $\tau$, MOIRAI processes all
channels in the local temporal window and STCPE aggregates the overlapping window outputs:

\begin{equation}
\label{eq:stcpe}
\begin{aligned}
\mathbf{H}_{\tau}
&=
\mathrm{MOIRAI}\!\left(
    \tilde{\mathbf{X}}_{[:,\, \tau-m : \tau+m, :]}
\right)
\in \mathbb{R}^{C \times w \times (d_{\mathrm{model}}/8)}, \\
\mathbf{E}_{\mathrm{STCPE}}[:, t, :]
&=
P_{\uparrow}
\!\left(
\sum_{\tau:\, |t-\tau|\le m}
\mathbf{H}_{\tau}[:,\, t-\tau+m, :]
\right)
\in \mathbb{R}^{C \times d_{\mathrm{model}}},
\qquad t=1,\dots,N .
\end{aligned}
\end{equation}

Here, $t-\tau+m$ indexes the relative temporal offset within the window, and boundary handling is omitted
for clarity. STCPE therefore preserves temporal translation equivariance through sliding windows and channel
permutation equivariance through MOIRAI, allowing DIVER-1 to encode local spatio-temporal context without
learning biases tied to a fixed channel order.

\paragraph{Position, modality, and spectral embeddings.}
We augment each patch with channel metadata and frequency information. Channel metadata is encoded as 
$[\mathbf{E}_{\text{position}}, \mathbf{E}_{\text{modality}}]=[PE^{(x)}(x), PE^{(y)}(y), PE^{(z)}(z), \mathbf{e}_{\text{type}}]$, 
where $(x,y,z)$ are MNI coordinates, $\mathbf{e}_{\text{type}}$ denotes electrode type 
(depth for SEEG; grid/strip for ECoG), and $PE$ is the sinusoidal coordinate encoding from PopT~\citep{chau2025popt}. 
Unavailable coordinates are encoded with $\mathbf{E}_{\text{position}}=\mathbf{0}$. 
We also add $\mathbf{E}_{\text{spectral}}$, obtained by FFT-transforming each patch and linearly projecting the features to $d_{\text{model}}$, following CBraMod~\citep{wang2024cbramod}.

\paragraph{Any-variate Transformer encoder (MOIRAI).}

The enhanced token grid $\mathbf{X}\in\mathbb{R}^{C\times N\times d_{\mathrm{model}}}$ is processed by any-variate Transformer blocks adapted from MOIRAI~\citep{woo2024moirai}. Each token is indexed by temporal patch index $i$ and channel index $m$. \textbf{The key design choice is any-variate attention}: instead of restricting attention to temporal-only, channel-only, or axis-factorized interactions, DIVER-1 applies full self-attention over all $CN$ electrode--time tokens. This directly addresses the restricted-attention bottleneck while preserving compatibility with variable iEEG inputs, allowing each channel--time segment to interact with every other segment without assuming a fixed electrode layout.

To make full attention expressive rather than layout-agnostic, DIVER-1 directly embeds spatio-temporal structure into the attention computation itself. Specifically, any-variate attention combines RoPE for temporal offsets with a binary same-channel versus cross-channel attention bias. Omitting layer indices, head indices, and the usual scaling factor, the attention energy and output are

\begin{equation}
\label{eq:anyvariate_attention}
\begin{aligned}
E_{ij,mn} &= (\mathbf{W}^{Q}\mathbf{x}_{i,m})^{\top}\mathbf{R}_{i-j}(\mathbf{W}^{K}\mathbf{x}_{j,n}) + u^{(1)}\mathbbm{1}_{\{m=n\}} + u^{(2)}\mathbbm{1}_{\{m\neq n\}}, \\
\mathbf{z}_{i,m} &= \sum_{j=1}^{N}\sum_{n=1}^{C}
\frac{\exp(E_{ij,mn})}{\sum_{k=1}^{N}\sum_{o=1}^{C}\exp(E_{ik,mo})}
\mathbf{W}^{V}\mathbf{x}_{j,n}.
\end{aligned}
\end{equation}

Here, $\mathbf{W}^{Q}$, $\mathbf{W}^{K}$, and $\mathbf{W}^{V}$ are the attention projections, $\mathbf{R}_{i-j}$ is the RoPE matrix for temporal offset $i-j$, and $u^{(1)},u^{(2)}\in\mathbb{R}$ are learnable same-channel and cross-channel biases that may differ by layer and attention head. Because the binary bias depends only on whether two tokens originate from the same channel, rather than on the absolute channel identity, the encoder preserves channel permutation equivariance while still distinguishing within-electrode dynamics from cross-electrode interactions. Thus, even when electrode metadata are unavailable, DIVER-1 remains channel-relation aware.

\subsection{Pretraining DIVER-1}

DIVER-1 is pretrained with masked multi-domain reconstruction on randomly sampled spatio-temporal subviews of 30 s iEEG segment. 

\paragraph{Spatio-Temporal Resampling (STR).}  To improve \textbf{robustness to variable electrode coverage and temporal context}, DIVER-1 is pretrained on random spatio-temporal subviews of each 30\,s iEEG segment. At each step, we sample \(C' \le \min(C,32)\) contacts and \(N' \le 30\) temporal patches from a scaled \(\mathrm{Beta}(3,1)\) distribution, favoring larger subviews. The temporal cap keeps the effective context budget comparable across DIVER-1-1s and DIVER-1-0.1s. Details in Appendix~\ref{appendix_subsec:str}.

\paragraph{Multi-Domain Reconstruction Objective (MDRO).}       Electrophysiological signals are commonly analyzed as time-domain waveforms, frequency spectra, and time-frequency representations, each capturing transient responses, rhythmic activity, and evolving spectral structure \citep{luck2014introduction,cohen2014analyzing,buzsaki2012origin}. Following these standard views and recent electrophysiology foundation models \citep{wang2023brainbert,zhang2023brant,jiang2024labram,wang2024cbramod}, DIVER-1 reconstructs multiple signal domains rather than only the raw waveform. Details in Appendix~\ref{appendix_subsec:diver_arch_pretext}.

\section{Experimental setup}

\paragraph{QA/QC and preprocessing.}

All iEEG recordings were high-pass filtered at 0.5~Hz, notch filtered at 60~Hz, resampled to 500~Hz, and segmented into 30-second windows. Clipping-based QA/QC removed electrodes when more than $3.33\%$ of samples exceeded the clipping threshold, and discarded segments when more than $50\%$ of channels were affected. After clipping, amplitudes were scaled from $[-200, 200]~\text{\textmu V}$ to $[-1, 1]$, following \citet{wang2024cbramod}. Further details are provided in Appendix~\ref{appendix_subsec:QAQC_preproc}.

\paragraph{Pretraining dataset and setup.}
DIVER-1 was pretrained on a multi-country, cross-center iEEG corpus spanning ECoG and sEEG, comprising \(5{,}310\) hours, \(352\)k channel-hours, and 37 subjects. This is \textbf{\(54\times\) more channel-hours than the BrainTreeBank corpus used to pretrain BrainBERT, PopT, and BaRISTA}. DIVER-1 also exceeds Brant's 281k channel-hour private SEEG corpus, with an even larger gap in effective training tokens: Brant's 6 s patches yield 169M channel-patch tokens, whereas DIVER-1 yields 1.27B tokens for DIVER-1-1s and 12.7B tokens for DIVER-1-0.1s. Thus, DIVER-1 exposes the model to $\mathbf{7.5\times}$ \textbf{and} $\mathbf{75\times}$ \textbf{more training tokens} at 1 s and 0.1 s resolution, respectively. This finer temporal granularity is important for subsecond cognitive decoding while preserving coverage of longer-timescale clinical dynamics. Dataset details are summarized in Table~\ref{tab:pretraining_dataset}; comparisons with prior iEEG foundation models are provided in Table~\ref{tab:previous_EFM}. 

At this scale, DIVER-1-0.1s and DIVER-1-1s were trained across model widths, yielding variants from 13M up to 1.83B parameters (Table~\ref{tab:model_params_pflops}). Since training the largest 1.83B-parameter variant required up to 128 A100 GPUs, naive hyperparameter sweeps at full scale are computationally prohibitive. We therefore adopt maximal update parameterization ($\mu$P)~\citep{muparamyang2022tensor}, which preserves update magnitudes across model widths and enables $\mu$Transfer: hyperparameters tuned on small proxy models transfer directly to larger ones, letting us scan DIVER-1 from 13M to 1.83B parameters under a single configuration. Details are in Appendix~\ref{appendix_subsec:pretraining_setup} and~\ref{appendix_subsec:mu_param}.

\begin{figure}
    \centering
    \includegraphics[width=0.9\linewidth]{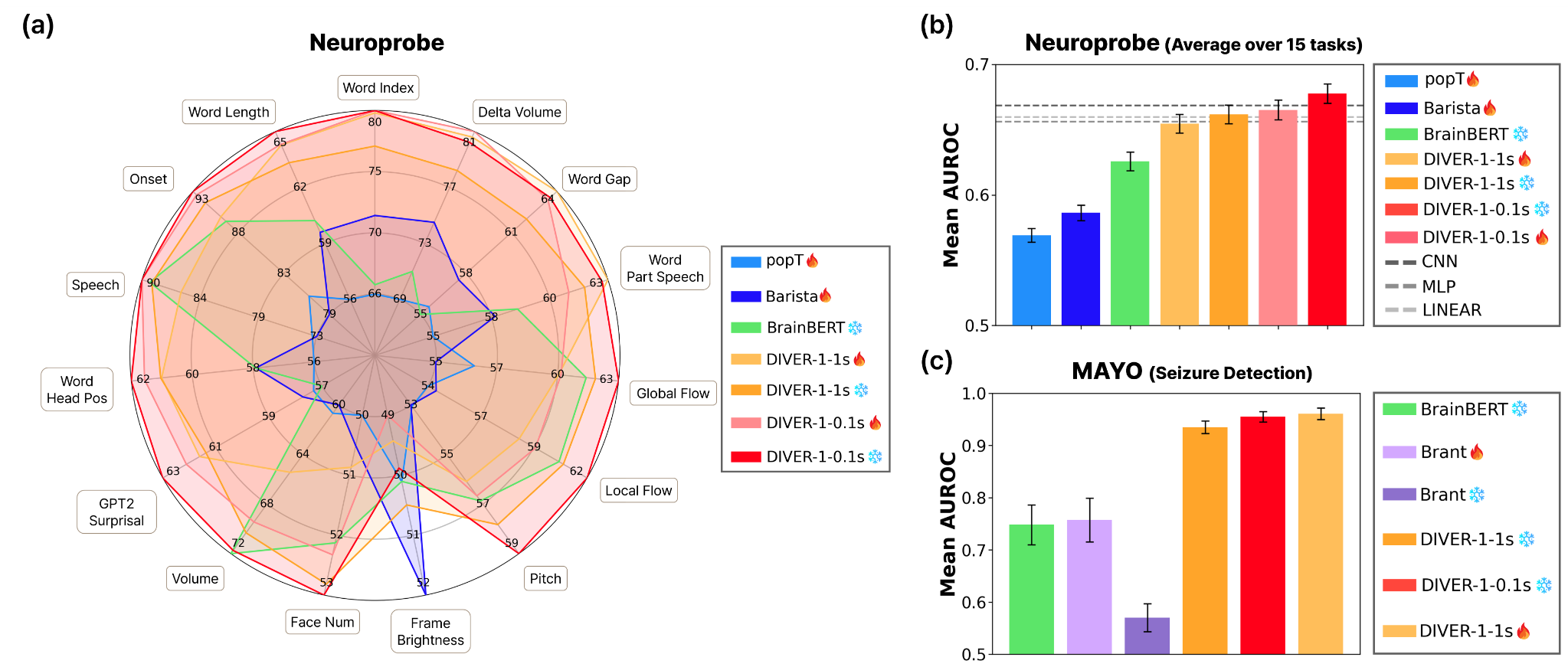}
    \caption{\textbf{DIVER-1 achieves strong iEEG decoding across naturalistic and clinical benchmarks.}
(a) AUROC across the 15 Neuroprobe tasks using 1-s samples recorded during movie watching. DIVER-1-0.1s achieves state-of-the-art or competitive performance on most tasks compared with prior iEEG foundation models. 
(b) Mean AUROC across the 15 Neuroprobe tasks. DIVER-1-0.1s performs best with linear probing. 
(c) AUROC on MAYO seizure detection benchmark using 6~s windows. DIVER-1-1s achieves the strongest seizure detection performance.
Snowflake denotes frozen-encoder linear probing, and fire icons denote full finetuning. Error bars denote SEM across all folds. Brant could not be evaluated on Neuroprobe due to its long patch size (6 s), and PopT and BaRISTA could also not be evaluated on MAYO due to missing channel coordinates.}
    \label{fig:SOTAperf}
\end{figure}
\paragraph{Evaluation datasets and tasks.}

We evaluate DIVER-1 on the broadest iEEG downstream suite among the compared foundation-model baselines, spanning naturalistic cognitive decoding and clinical seizure detection. For naturalistic decoding, we use Neuroprobe~\citep{zahorodnii2025neuroprobe}, a 15-task benchmark built on BrainTreebank~\citep{wang2024brain} movie-watching iEEG that extends the four probes used in prior BrainBERT, PopT, and BaRISTA evaluations. For clinical
evaluation, we use the MAYO seizure-detection dataset~\citep{bbrinkm2014upenn}. Dataset details are provided in Appendix~\ref{appendix_subsec:finetuning_dataset_description}.

\paragraph{Baselines and finetuning.}
We compare DIVER-1 against four self-supervised iEEG foundation models---BrainBERT~\citep{wang2023brainbert}, PopT~\citep{chau2025popt}, Brant~\citep{zhang2023brant}, and BaRISTA~\citep{oganesian2025barista}---as well as benchmark-reported supervised baselines when available. These models are evaluated with linear probing or full finetuning using a linear classifier on flattened token representations. Details are provided in Appendix~\ref{appendix finetuning setup}.

\section{Results and Discussion}

\subsection{DIVER-1 learns transferable iEEG representations}

\paragraph{DIVER-1 achieves state-of-the-art iEEG decoding under cross-dataset shift.}

Figure~\ref{fig:SOTAperf} evaluates DIVER-1 on Neuroprobe, a 15-task naturalistic iEEG decoding benchmark using 1~s windows, and MAYO, a seizure detection benchmark using 6~s windows. On Neuroprobe, \textbf{DIVER-1-0.1s achieves the best mean AUROC among all evaluated pretrained iEEG models and supervised model baselines} (Figure~\ref{fig:SOTAperf}b). Importantly, under this protocol, prior evaluated iEEG foundation models underperform the simple linear spectrogram baseline in mean AUROC, whereas \textbf{DIVER-1-0.1s is the first iEEG foundation model to surpass it}. At the task level, DIVER-1-0.1s is state-of-the-art or competitive on most tasks (Figure~\ref{fig:SOTAperf}a), except Frame Brightness, where DIVER-1 and most baselines remain close to chance, suggesting weak task decodability rather than model-specific failure. Notably, \textbf{linear probing is highly competitive}: the frozen DIVER-1-0.1s outperforms its full-finetuning counterpart on Neuroprobe, indicating that pretrained representations are already linearly useful for naturalistic decoding. On MAYO, \textbf{DIVER-1-1s achieves the strongest seizure detection performance}, substantially outperforming BrainBERT and Brant (Figure~\ref{fig:SOTAperf}c), with linear probing remaining nearly on par with full finetuning. The different best-performing variants are consistent with the evaluation timescales: DIVER-1-0.1s is better matched to fine-grained 1~s naturalistic decoding, whereas DIVER-1-1s is better matched to the longer 6~s clinical seizure detection context. 

These results are notable because DIVER-1 achieves them under a stricter \textbf{cross-center, cross-population pretraining--finetuning shift} than prior Neuroprobe baselines. BrainBERT, PopT, and BaRISTA are pretrained on BrainTreeBank, the same corpus used to construct Neuroprobe. In contrast, \textbf{DIVER-1 is pretrained on external adult iEEG recordings from different recording centers}, spanning both SEEG and ECoG, while Neuroprobe evaluates pediatric SEEG. We further disentangle architecture and pretraining data in Table~\ref{tab:ablation_dataset_arch}. Finally, despite having only 13M parameters, DIVER-1-0.1s outperforms larger prior models such as BrainBERT and PopT; Section~\ref{subsec:scalinglaw_dataconstrained} analyzes this size--performance relationship and shows that \textbf{broader heterogeneous data and sufficient training duration matter more than parameter count alone}.

\begin{table}
\centering
\caption{\textbf{Architecture and pretraining strategy ablation.} Single-component ablations of DIVER-1-0.1s on representative Neuroprobe tasks. Rows remove components from three groups: encoder attention, input embeddings, and pretraining strategy. All models are $d_\text{model}=256$ and pretrained for 8 epochs. Results are reported as mean AUROC $\pm$ SEM across subjects, trials, and folds.}
\scriptsize
\color{black}{
\begin{tabular}{llcccc}
    & & speech & onset & volume & pitch \\ 
\midrule
\multirow{3}{*}{\rotatebox[origin=c]{90}{\textbf{Atten.}}}
 & w.o. RoPE & $0.886 \pm 0.013$ & $0.916 \pm 0.011$ & $0.693\pm 0.019$ & $0.579\pm 0.008$ \\
 & w.o anyV attention & $0.889 \pm 0.013$ & $0.919 \pm 0.009$ & $0.699 \pm 0.018$ & $0.579 \pm 0.007$ \\
 & w.o RoPE and anyV attention (Vanilla Attention) & $0.870 \pm 0.014$ & $0.898 \pm 0.012$ & $0.669 \pm 0.018$ & $0.560 \pm 0.006$ \\
\midrule 
\multirow{4}{*}{\rotatebox[origin=c]{90}{\textbf{Embedding}}}
 & w.o. STCPE & $0.879 \pm 0.014$ & $0.911 \pm 0.013$ & $0.686 \pm 0.019$ & $0.572 \pm 0.007$ \\
 & w.o Channel sub-modality emb. & $0.892 \pm 0.011$ & $0.919 \pm 0.009$ & $0.690 \pm 0.017$ & $0.577 \pm 0.006$ \\
 & w.o Channel 3d position emb. & $\mathbf{0.900 \pm 0.011}$ & $\mathbf{0.927\pm 0.009}$ & $\mathbf{0.710 \pm 0.018}$ & $\mathbf{0.584 \pm 0.009}$ \\
 & w.o Spectral feature emb. & ${0.885 \pm 0.013}$ & $0.919\pm 0.010$ & $0.694 \pm 0.019$ & $0.571 \pm 0.010$ \\
\midrule 
\multirow{6}{*}{\rotatebox[origin=c]{90}{\textbf{Pretraining.}}}
 & w.o Multi-domain reconstruction (only raw) & $0.875 \pm0.014$ & $0.916 \pm0.010$ & $0.680 \pm 0.017$ & $0.569 \pm 0.006$ \\ 
 & w.o channel and time resampling & $0.807 \pm0.017$ & $0.884 \pm0.011$ & $0.628 \pm 0.015$ & $0.552 \pm 0.006$ \\ 
 & w.o channel resampling & $0.856 \pm 0.016$ & $0.895 \pm 0.012$ & $0.647 \pm 0.016$ & $0.553 \pm 0.006$ \\ 
 & w.o time resampling & $0.854 \pm 0.016$ & $0.900 \pm 0.011$ & $0.652 \pm 0.016$ & $0.558 \pm 0.005$ \\ 
 & w.o channel and time resampling (1sec window) & $0.878 \pm 0.014$ & $0.902 \pm 0.012$ & $0.662 \pm 0.016$ & $0.559 \pm 0.006$ \\ 
 & resampling but 1sec window & $0.862 \pm 0.015$ & $0.902 \pm 0.012$ & $0.661 \pm 0.016$ & $0.561 \pm 0.006$ \\ 
\midrule
\multicolumn{2}{l}{DIVER-1-0.1s (resampling at 3sec window)} & $0.890 \pm 0.013$ & $0.922 \pm 0.009$ & $0.698 \pm 0.018$ & $0.572 \pm 0.008$ \\ 
\bottomrule
\end{tabular}
}
\label{table:archablation}
\end{table}

\begin{table}[!htbp]
\centering
\caption{\textbf{Ablation of absolute 3D channel-position embedding on models pretrained on the BTB dataset} When pretraining and downstream distributions are matched, the position embedding improves performance across all tasks unlike Table~\ref{table:archablation}.}
\scriptsize
\begin{tabular}{lccccc}
Data & Model & Speech & Onset & Volume & Pitch \\
\midrule
BTB & w.o. channel 3d position emb. & $0.756 \pm 0.023$ & $0.827 \pm 0.018$ & $0.600 \pm 0.024$ & $0.539 \pm 0.011$ \\
BTB & DIVER-1-0.1s & $\bm{0.770 \pm 0.028}$ & $\bm{0.859 \pm 0.018}$ & $\bm{0.639 \pm 0.023}$ & $\bm{0.558 \pm 0.010}$ \\
\bottomrule
\end{tabular}
\label{tab:btb_pos_ablation}
\end{table}

\paragraph{Architecture and pretraining strategy ablation.}

Table~\ref{table:archablation} reports component-wise ablations of DIVER-1-0.1s. 
The largest changes come from the resampling ablations. Replacing DIVER-1's attention with vanilla self-attention consistently reduces performance, while removing RoPE or the any-variate same/cross-channel bias alone has smaller, task-dependent effects, suggesting complementary benefits. Among embeddings and objectives, STCPE and spectral features provide modest gains, channel sub-modality has limited impact, and MDRO improves over raw-only reconstruction.

The resampling ablations show that simply matching the downstream 1~s window during pretraining is insufficient. Specifically, our default 3~s STR pretraining outperforms both 1~s pretraining without resampling and 1~s pretraining with STR, indicating that longer-context pretraining transfers to 1~s decoding, in line with works in LLMs~\cite{gao-etal-2025-trainlongeffective,zheng-etal-2025-longhelpsshort}. Within the 1~s setting, STR has only a modest effect on downstream performance, suggesting that stochastic resampling improves input flexibility without substantially compromising task-relevant representations.

The absolute 3D channel-position embedding shows a different pattern: removing it improves all four tasks in Table~\ref{table:archablation}. We hypothesize this reflects pretraining--downstream distribution shift. DIVER-1 is pretrained on adult population data, whereas Neuroprobe evaluates pediatric subjects, so age-related differences in brain geometry and electrode localization may reduce the transferability of absolute coordinates. In fact, when pretraining and downstream evaluation are distribution-matched on BrainTreeBank, 3D position embedding improves every task (Table~\ref{tab:btb_pos_ablation}). Thus, \textbf{absolute coordinates help when electrode geometries align, but can hurt under cross-age or cross-population transfer}.

\begin{table}[!htbp]
\centering
\caption{\textbf{Ablation of architectural and data-scale effects.} Models are pretrained on the same BTB corpus to isolate architectural effects and data scaling effect of DIVER. Results are mean AUROC $\pm$ SEM across subjects, trials on single fold (fold0). DIVER-1-1s is pretrained for 32 epochs.}
\label{tab:ablation_dataset_arch}
\scriptsize
\begin{tabular}{llccc}
\toprule
\textbf{Model} 
& \textbf{Pretraining Dataset} 
& \textbf{Speech} 
& \textbf{Onset} \\
\midrule
BrainBERT (frozen) & BrainTreeBank  & $0.606 \pm 0.021$ & $0.754 \pm 0.027$ \\
PopT               & BrainTreeBank   & $0.657 \pm 0.024$ & $0.648 \pm 0.029$ \\
BaRISTA            & BrainTreeBank  & $0.678 \pm 0.041$ & $0.745 \pm 0.038$ \\
DIVER-1-1s              & BrainTreeBank  & $0.770 \pm 0.028$ & $0.859 \pm 0.018$ \\
\midrule
DIVER-1-1s            & Private 1.1\% (x1.0 BTB)  & $0.710 \pm 0.015$ & $0.758 \pm 0.016$ \\
DIVER-1-1s          & Private 17.7\% (x15.8 BTB)   & $0.840 \pm 0.022$ & $0.898 \pm 0.014$ \\
DIVER-1-1s           & Private 70.6\%  (x63.3 BTB)   & $0.847 \pm 0.020$ & $0.905 \pm 0.013$ \\
\bottomrule
\end{tabular}
\end{table}

\paragraph{Disentangling architectural and data-scale contributions.}
Because DIVER-1 combines a new architecture with substantially larger pretraining data, we run two controlled comparisons to separate architecture from data scale (Table~\ref{tab:ablation_dataset_arch}). First, under matched BrainTreebank (BTB) pretraining, \textbf{DIVER-1-1s outperforms prior architectures} on speech and onset detection, indicating that its gains are not due solely to larger pretraining data. Second, with the DIVER-1-1s architecture fixed, we pretrain on size-controlled external iEEG corpora from a different country and age group. A BTB-sized external corpus underperforms BTB pretraining, showing that distributional alignment matters at limited scale; however, increasing the external corpus to 15.8$\times$ and 63.3$\times$ BTB scale closes and reverses this gap. Thus, \textbf{architecture improves performance under matched data, while sufficient pretraining scale can overcome demographic and geographic mismatch}.

\subsection{Scaling laws reveal a data-constrained iEEG regime}
\label{subsec:scalinglaw_dataconstrained}

Table~\ref{tab:ablation_dataset_arch} suggests that DIVER-1 benefits from large-scale pretraining. We therefore sweep training compute, data tokens, model size, training duration, and subject diversity to ask whether self-supervised iEEG pretraining scales predictably, and which axes matter most for downstream transfer.

\begin{figure}[!ht]
    \centering
    \includegraphics[width=0.9\linewidth]{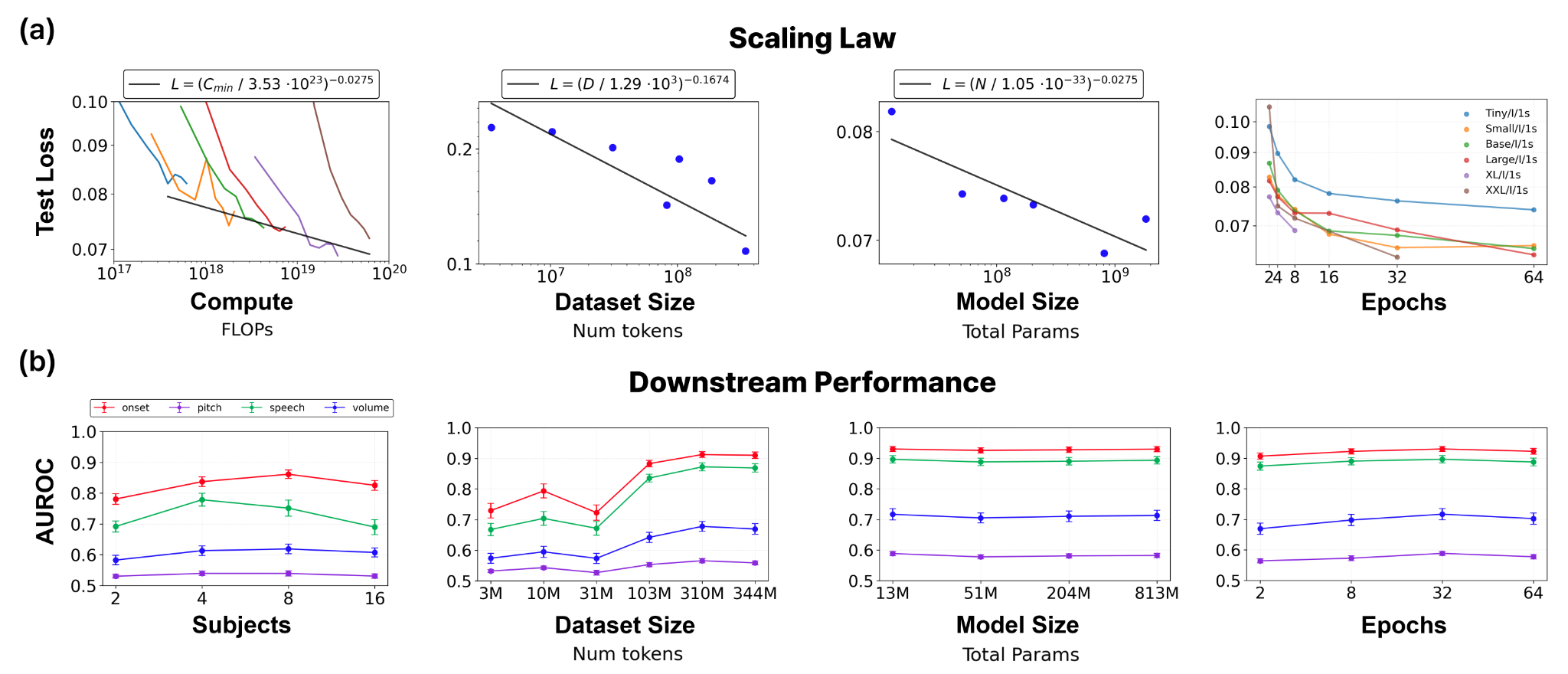}
        \caption{\textbf{Scaling laws and downstream scaling of DIVER-1.} 
    (a) Held-out pretraining loss of DIVER-1-1s follows \citet{kaplan2020scaling}-style power-law trends as training compute, data, and model parameters increase. Training-duration curves show that larger parameter-count variants require sufficient repeated exposure to the data before outperforming smaller variants, consistent with data-constrained scaling \citep{data_constrained_scaling_muennighoff2023scaling}. See Appendix~\ref{appendix_subsec:extended_kaplan_ieeg} for DIVER-1-0.1s and extended results.
    (b) Downstream AUROC on four Neuroprobe tasks under controlled scaling of subject count, dataset size, model size, and training duration on DIVER-1-0.1s. Dataset size and training duration yield most consistent gains, whereas model size and subject diversity show mixed effects under fixed data budgets.}

    \label{fig:Scaling}
\end{figure}

\textbf{Scaling-law setup.}     Because iEEG data are scarce and multi-pass training is often unavoidable, we fit the data-constrained scaling law of \citet{data_constrained_scaling_muennighoff2023scaling}, which extends Kaplan-style power laws~\citep{kaplan2020scaling} by replacing raw data and parameter counts with effective counts. Specifically, we model loss as \(L(N,D)=A(N')^{-\alpha}+B(D')^{-\beta}+E\), with \(D'=g(U_D,R_D,R_D^*)\), \(N'=g(U_N,R_N,R_N^*)\), and \(g(U,R,R^*)=U[1+R^*(1-e^{-R/R^*})]\). Here, \(U_D\) and \(R_D\) denote unique tokens and additional passes over them, while \(R_D^*\) controls diminishing returns from repetition; \(U_N,R_N,R_N^*\) play the analogous role for excess model capacity. This lets us test whether iEEG pretraining scales predictably across data, model size, and training duration. Details can be found in Appendix \ref{appendix_subsec:neural_scaling_law}.

\textbf{Fixed-duration sweeps show predictable pretraining scaling.}     At fixed training duration, held-out loss decreases approximately linearly on log--log axes as compute, data size, and model size increase (Figure~\ref{fig:Scaling}a, right), consistent with Kaplan-style scaling~\citep{kaplan2020scaling} and suggesting that \textbf{masked iEEG reconstruction scales predictably}. Further training further reduce loss, but larger models only outperform smaller ones after sufficient iterations; at short durations, they can remain undertrained. Thus, \textbf{model capacity helps only when supported by sufficient training duration}, motivating compute-optimal allocation analysis below.

\textbf{Downstream scaling favors data and training duration over parameter count.}        Figure~\ref{fig:Scaling}(b) shows that increasing dataset size gives the most reliable downstream gains across representative Neuroprobe tasks, whereas increasing model size produces weaker and more mixed effects, in line with \citet{anonymous2026are}. This pattern is also reflected in the pretraining scaling curves in Figure~\ref{fig:Scaling}(a): the dataset-size axis produces a larger reduction in held-out reconstruction loss than the model-size axis. Together, these results suggest that DIVER-1 is currently more data-limited than parameter-limited.

Training duration is another reliable scaling axis: AUROC generally improves up to 32 training passes, with mild decline at 64. Subject diversity is non-monotonic under fixed total pretraining volume, suggesting an optimal balance between the number of subjects and volume per subject. Together, these results motivate the compute-optimal analysis below: \textbf{for iEEG foundation models, unique data and sufficient training duration should be prioritized before scaling parameter count}.

\begin{figure}
    \centering
    \includegraphics[width=1.0\linewidth]{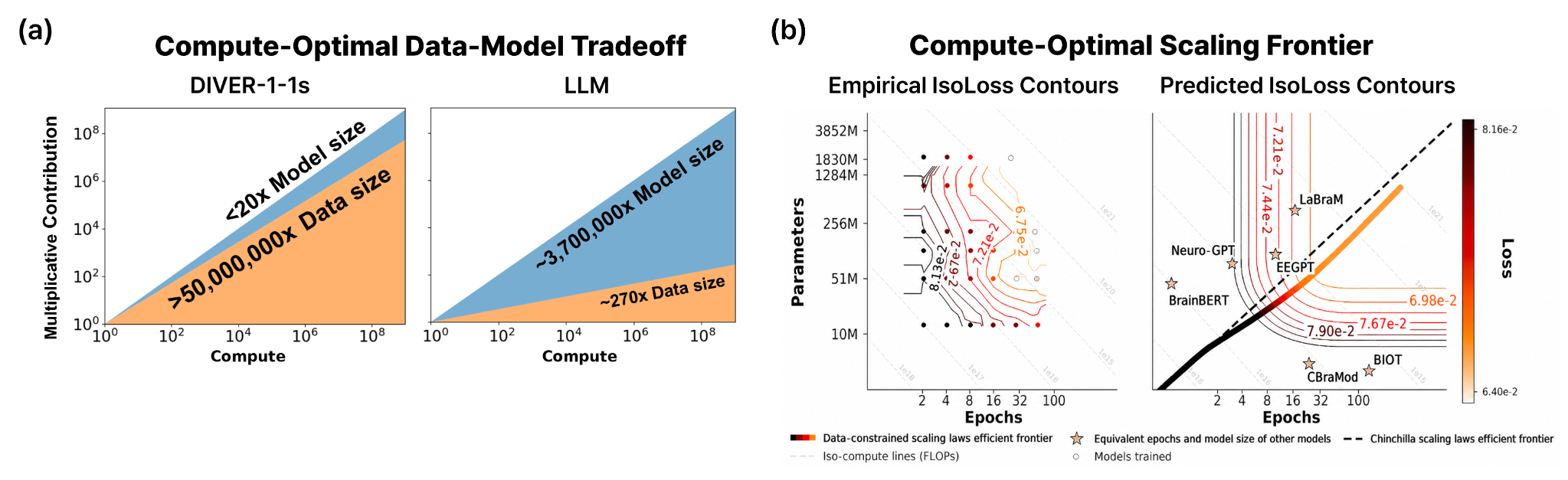}
    \caption{\textbf{Compute-optimal scaling of DIVER-1.}
    (a) \textbf{Optimal data--model allocation when training duration is fixed.}
    With training fixed to two passes, we use the fitted DIVER-1-1s scaling law to estimate how the compute-optimal dataset size and parameter count should grow as total compute increases. The y-axis shows the fold-increase assigned to each factor. Compared with LLM scaling of~\citet{kaplan2020scaling}, DIVER-1-1s allocates much more of the compute increase to data growth and much less to parameter growth.
    (b) \textbf{Optimal parameter--training-duration allocation when the corpus is fixed.}
    Using fixed pretraining corpus, we plot held-out loss over model size and number of training passes. IsoCompute curves mark equal FLOPs, and IsoLoss contours show equal reconstruction loss. The efficient frontier favors small-to-mid-sized models trained longer over largest models trained briefly. \textcolor{black}{See Appendix~\ref{appendix_subsec:data_constrained_ieeg}} for result of DIVER-1-0.1s variant.}
    \label{fig:ComputeOptimal}
\end{figure}

\paragraph{Compute-optimal allocation.}\label{result:compute_optimal}

The scaling-law fits give two practical compute-allocation rules for iEEG. Classical compute-optimal analyses ask how to split compute between data and parameters when fresh data can grow with compute~\citep{kaplan2020scaling,hoffmann2022chinchilla}. In iEEG, however, data is usually fixed, so extra compute can only be spent on larger models or more training passes over the same recordings.

\textbf{First, scale data before parameters when new recordings are available.} Figure~\ref{fig:ComputeOptimal}(a) shows that at an epoch of two, DIVER-1-1s favors a much more data heavy allocation than LLMs~\citep{kaplan2020scaling}. DIVER-1-0.1s shows the same trend (Appendix~\ref{appendix_subsec:extended_kaplan_ieeg}), indicating that better iEEG foundation models depend primarily on expanding unique pretraining recordings, not simply adding parameters.

\textbf{Second, when the corpus is fixed, scale training duration before model size.} Figure~\ref{fig:ComputeOptimal}(b) shows that realistic compute budgets favor small-to-mid-sized models trained for more passes, rather than very large models trained briefly. The predicted contours from Eq.~\ref{eq:dataconstrainscaling} reproduce this frontier, suggesting that the fitted law can guide parameter--duration choices before expensive pretraining. Overall, the compute-efficient order is: \textbf{increase unique data first, train sufficiently long second, and scale parameters last}. Fitted parameters and additional plots are provided in Appendix~\ref{appendix_subsec:data_constrained_ieeg}.

\section{Conclusion}

We introduced DIVER-1, a self-supervised iEEG foundation model designed to learn transferable representations from heterogeneous clinical recordings. By combining an iEEG-specific architecture with large-scale external pretraining, DIVER-1 improves cross-dataset transfer on both naturalistic cognitive decoding and clinical seizure detection, surpassing prior iEEG foundation models as well as non-linear baselines. Our scaling study further shows that iEEG pretraining remains data-constrained: pretraining data and longer training are more reliable drivers of performance than parameter growth alone. These results suggest a practical path for future iEEG foundation models: scale data first, train longer second, and scale model size only after those bottlenecks are addressed.


\appendix
\newpage
\section{Appendix}
\setcounter{section}{0}  
\IfFileExists{appendix.tex}{
\section*{Appendix Table of Contents}
\noindent\rule{\textwidth}{1pt}
\vspace{0.5em}

\noindent\textbf{Section \ref{appendix:related_works}:} Related Works \\
\hspace*{2em}\ref{appendix_subsec:ephys_data_found}   \ Electrophysiology Data Foundation Models \\
\hspace*{2em}\ref{appendix_subsec:neural_scaling_law}  \ Neural Scaling Law

\noindent\textbf{Section \ref{appendix:experimental_setup_detail}:} Experimental Setup Details \\
\hspace*{2em}\ref{appendix_subsec:tested_models}  \ Tested Models \\
\hspace*{2em}\ref{appendix_subsec:pretraining_dataset}  \ Pretraining Dataset \\
\hspace*{2em}\ref{appendix_subsec:downstream_task}  \ Downstream Task and Dataset Overview \\
\hspace*{2em}\ref{appendix_subsec:pretraining_setup}  \ Pretraining Setup and Model Scaling \\
\hspace*{2em}\ref{appendix_subsec:str}  \ Spatio-Temporal Resampling \\
\hspace*{2em}\ref{appendix_subsec:diver_arch_pretext}  \ DIVER Architecture and Pretext Task Hyperparameter Setting \\
\hspace*{2em}\ref{appendix_subsec:pretraining_hyperparam}  \ Pretraining Hyperparameter Search using $\mu$-Parameterization and $\mu$-Transfer \\
\hspace*{2em}\ref{appendix_subsec:mu_param}  \ $\mu$-parameterization ($\mu P$) \\
\hspace*{2em}\ref{appendix_subsec:finetuning_setup}  \ Finetuning Setup

\noindent\textbf{Section \ref{Appendix:Scaling Law}:} Scaling Law \\
\hspace*{2em}\ref{appendix_subsec:scaling_law_exp}  \ Scaling Law Experiment Details \\
\hspace*{2em}\ref{appendix_subsec:extended_kaplan_ieeg}  \ Extended Kaplan Scaling Law Results \\
\hspace*{2em}\ref{appendix_subsec:data_constrained_ieeg}  \ Data-Constrained Scaling Law Results \\

\noindent\textbf{Section \ref{appendix_sec:extended_results}:} Extended Results \\
\hspace*{2em}\ref{appendix_subsec:comprehensive_diver_downstream}  \ Comprehensive DIVER Downstream Task Results \\
\hspace*{2em}\ref{appendix_subsec:perf_eval_across_diver1}  \ Performance Evaluation across DIVER-1 Model Configurations \\
\hspace*{2em}\ref{appendix_subsec:interpretation_results}  \ Interpretation Results

\noindent\textbf{Section: \ref{appendix:data_details}} Data Details \\
\hspace*{2em}\ref{appendix_subsec:pretraining_dataset_description}  \ Pretraining Dataset Description \\
\hspace*{2em}\ref{appendix_subsec:finetuning_dataset_description}  \ Finetuning Dataset Description \\
\hspace*{2em}\ref{appendix_subsec:QAQC_preproc}  \ QAQC and Preprocessing

\noindent\textbf{Section \ref{appendix:previous_EFM}:} Comparison With Existing EEG/iEEG Foundation Models

\vspace{0.5em}
\noindent\rule{\textwidth}{1pt}
\newpage

\section{Related Works}\label{appendix:related_works}

\subsection{Electrophysiology Data Foundation Models}\label{appendix_subsec:ephys_data_found} 
Ephys decoding has progressed from pipelines that coupled hand-crafted features with classical classifiers to end-to-end deep architectures that learn task-relevant representations directly from raw signals. Early EEG studies relied on feature engineering (e.g., band-power, CSP) paired with SVMs or LDA; subsequent work introduced convolutional backbones and sequence models that absorb spectral–temporal patterns with minimal preprocessing. Building on advances in self-supervised learning (SSL) and “foundation” paradigms, recent efforts pretrain large models on heterogeneous, weakly labeled or unlabeled EEG/iEEG corpora and adapt them to diverse downstream tasks (e.g., event detection, cognitive state decoding, BCI control). Representative lines include large-scale EEG transformers and montage-aware encoders (e.g., EEGFormer\citep{chen2024eegformer}, NEURO-GPT\citep{cui2024neurogpt}, LaBraM\citep{jiang2024labram}), intracranial representation learners (e.g., BrainBERT\citep{wang2023brainbert}, foundation models for intracranial neural signals), and broader neuro-sequence backbones emphasizing transfer and robustness (e.g., CBraMod\citep{wang2024cbramod}).

\subsection{Neural Scaling Law}\label{appendix_subsec:neural_scaling_law}
\textbf{Kaplan \citep{kaplan2020scaling} scaling law} provides a principled framework for predicting model performance as a function of model size and dataset size for large language models. They demonstrated that language model cross-entropy loss follows smooth power-law relationships with respect to model parameters ($N$) and training data ($D$). Concretely, they proposed a relation of the form
\begin{align}
    L(N,D) &= \frac{A}{N^{\alpha}} + \frac{B}{D^{\beta}} + E
\end{align}
where $A,B,E$ are constants, and $\alpha,\beta > 0$ are scaling exponents. This formulation implies that increasing the number of parameters or training data yields a predictable reduction in loss, enabling systematic optimization of compute allocation across model size and training duration.

\textbf{Data-constrained scaling law.} In scientific domains, including Ephys, unique high-quality data are inherently limited. To address this, \citet{data_constrained_scaling_muennighoff2023scaling} extended the scaling framework to data-constrained regimes, where models must repeatedly train on the same corpus for multiple epochs. They proposed a modified law
\begin{align}
    L(N,D) &= \frac{A}{(N')^{\alpha}} + \frac{B}{(D')^{\beta}} + E
\end{align}\label{eq:dataconstrainscaling}
where the effective number of parameters $N'$ and effective tokens $D'$ account for diminishing returns: repeated tokens are progressively less valuable, and excessively large models are less sample-efficient. They further define
\begin{align}
    D' &= U_D + U_D R_D^{*}\Big(1 - e^{-\tfrac{R_D}{R_D^{*}}}\Big) \\
    N' &= U_N + U_N R_N^{*}\Big(1 - e^{-\tfrac{R_N}{R_N^{*}}}\Big)
\end{align}
where $U_D$ is the number of unique tokens, $R_D$ is the number of repetitions (epochs$-1$), and 
$R_D^{*}$ is a learned constant describing the “half-life" of repeated data. Analogously, $U_N$ is the compute-optimal parameter count for $U_D$, and $R_N^{*}$ governs diminishing returns beyond that point. Empirical results suggest that up to $\sim$4 epochs, repeated data is almost as useful as new data, but returns decrease sharply thereafter. 

Compared to the Chinchilla law \citep{hoffmann2022chinchilla}, which assumes abundant data and one training epoch, this formulation makes epoch count itself a central axis of scaling. This perspective is crucial for domains constrained by limited samples, such as iEEG, and it guides how compute should be allocated between model size and additional training passes.

Finally, the data-constrained scaling law also connects scaling analysis to more fundamental theory. Recent work has suggested that the exponents governing power-law scaling, such as $\alpha$ and $\beta$, are intimately connected to the \emph{intrinsic dimension} of the underlying data manifold~\citep{Sharma2022Scaling}. In particular, it has been argued that the data exponent $\beta$ can be interpreted in terms of an effective dimension $d$ via an approximate inverse relation of the form:
\begin{align}
    \beta \;\approx\; \frac{2}{d}.
\end{align}

From this view, estimating $\beta$ provides not only a measure of how data volume translates
into performance but also an insight into the intrinsic dimensionality of iEEG signals,
where scarcity and structural complexity are defining features.

\section{Experimental Setup Details}\label{appendix:experimental_setup_detail}
\subsection{Tested Models}\label{appendix_subsec:tested_models} 
We adopt a systematic naming convention for all DIVER model variants: $\textsc{DIVER-1-\text{Granularity}}_{\text{Size}}$.  The \textit{Granularity} indicates temporal resolution (0.1s or 1s window). The \textit{Size} component indicates model size (Tiny: 256, Small: 512, Base: 768, Large: 1024, XL: 2048, XXL: 3072 hidden dimensions). For example, $\textsc{DIVER-1-1s}_{\text{Base}}$ represents a base-sized model trained with 1s temporal patchsize. Unless otherwise specified, $\textsc{DIVER-1-1or0.1s}$ refers to the Tiny variant, i.e., $\textsc{DIVER-1-1or0.1s}_{\text{Tiny}}$.

\begin{table}[h]
\scriptsize
\centering
\caption{\textbf{Model configurations with measured parameters and  \textcolor{black}{total FLOPs per epoch.}}}
\label{tab:model_params_pflops}
\begin{tabular}{lccccc}
\toprule
\textbf{Models} 
& \textbf{\# Parameters} 
& \textbf{Modality} 
& \textbf{Hidden Dimension ($d_{\text{model}}$)} 
& \textbf{\textcolor{black}{total} FLOPs / epoch} \\
\midrule
$\textsc{DIVER-1-0.1s}_{\text{ Tiny}}$   & 12.72M   & 0.1s & 256  & \textcolor{black}{76.34{P}}   \\
$\textsc{DIVER-1-0.1s}_{\text{ Small}}$  & 50.75M  & 0.1s & 512  & \textcolor{black}{253.96{P}}    \\
$\textsc{DIVER-1-0.1s}_{\text{ Base}}$   & 114.07M  & 0.1s & 768  & \textcolor{black}{532.77{P}} \\
$\textsc{DIVER-1-0.1s}_{\text{ Large}}$  & 202.70M  & 0.1s & 1024 & \textcolor{black}{912.83{P}}  \\
$\textsc{DIVER-1-0.1s}_{\text{ XL}}$     & 810.19M  & 0.1s & 2048 & \textcolor{black}{3.40{E}} \\
$\textsc{DIVER-1-0.1s}_{\text{ XXL}}$    & 1.82B   & 0.1s & 3072 & \textcolor{black}{7.50{E}}  \\
\midrule
$\textsc{DIVER-1-1s}_{\text{ Tiny}}$     & 13.03M  & 1s   & 256  & \textcolor{black}{77.52{P}}  \\
$\textsc{DIVER-1-1s}_{\text{ Small}}$    & 51.36M   & 1s   & 512  & \textcolor{black}{256.44{P}}   \\
$\textsc{DIVER-1-1s}_{\text{ Base}}$     & 115.00M    & 1s   & 768  & \textcolor{black}{536.82{P}}  \\
$\textsc{DIVER-1-1s}_{\text{ Large}}$    & 203.95M  & 1s   & 1024 & \textcolor{black}{918.56{P}}  \\
$\textsc{DIVER-1-1s}_{\text{ XL}}$       & 812.85M  & 1s   & 2048 & \textcolor{black}{3.46{E}}  \\
$\textsc{DIVER-1-1s}_{\text{ XXL}}$      & 1.83B    & 1s   & 3072 & \textcolor{black}{7.64{E}} \\
\bottomrule
\end{tabular}
\end{table}

\subsection{Pretraining Dataset}\label{appendix_subsec:pretraining_dataset} 
DIVER-1 was pretrained on the largest iEEG corpus to date, comprising 352k channel-hours from 37 subjects using iEEG data (ECoG/sEEG).  The pretraining dataset description is given in Table~\ref{tab:pretraining_dataset}. Private data were collected from patients who provided informed consent to participate in an IRB-approved study.

\begin{table}
\scriptsize
\centering
\caption{\textcolor{black}{\textbf{Summary of DIVER-1 pretraining datasets.} }}
\label{tab:pretraining_dataset}
\begin{tabular}{lccccc}
\toprule
Datasets & Data Type & \# Subj. & Volume (channel-hours) & Duration (hours) & Sampling Rate (Hz) \\
\midrule
AJILE12~\citep{peterson2022ajile12} & ECoG & 12 & 124,423 & 1,282 & 1,000 \\
Self-collected iEEG (DIVER$_{\text{I}}$) & ECoG/sEEG & 25 & 227,612 & 4,028 & 2,000 \\
\midrule
\textbf{Total}  & --- & 37 & 352,035 & 5,310 & ---  \\
\bottomrule
\end{tabular}%
\end{table}

\subsection{Downstream Task and Dataset Overview}\label{appendix_subsec:downstream_task}

\textcolor{black}{Table~\ref{table:overview_downstream} provides a comprehensive overview of all downstream tasks and datasets used in our evaluation. Our evaluation covers diverse neural decoding objectives across visual, auditory, language domains and seizure detection.}

\textcolor{black}{\textbf{Neuroprobe} We evaluated on 15 tasks from the Neuroprobe (LITE) benchmark \citep{zahorodnii2025neuroprobe}, including visual perception (frame brightness, optical flow, face detection), auditory processing (volume, pitch, delta volume), and language processing (speech decoding, word prediction, onset detection, part-of-speech tagging). The Neuroprobe dataset contains depth electrode recordings from 6 subjects with 109-120 channels per subject, originally sampled at 2048Hz and was resampled to 500Hz to match our pretraining configuration.}

\textcolor{black}{\textbf{MAYO} For seizure detection, we evaluated the full foundation models on a dataset derived from MAYO Kaggle challenge dataset~\cite{bbrinkm2014upenn}. To ensure compatibility with Brant, witch has 6~s patch size, we concatenated the original 1~s samples in temporal order into 6~s segments and redefined the train/test splits.}

\begin{table}
\centering
\caption{\textcolor{black}{\textbf{Overview of downstream tasks and datasets.} Sampling rates were adjusted to 500Hz across all datasets to match the pretraining configuration. Arrows ($\to$) indicate resampling or channel selection from the original dataset.}}
\label{table:overview_downstream}
\scriptsize
\color{black}
\begin{tabular}{llccccc}
\toprule
\textbf{Modality} & \textbf{Task Name} & \textbf{Datasets} & \textbf{Sampling Rate} & \textbf{\# Ch.} & \textbf{\# Subj.} & \textbf{Label} \\
\midrule
\multirow{14}{*}{\textbf{SEEG}} 
  & frame\_brightness (visual) & \multirow{15}{*}{\shortstack{Neuroprobe\\(LITE)}} & \multirow{15}{*}{2048 $\to$ 500Hz} & \multirow{15}{*}{\shortstack{Var.\\(109--120)}} & \multirow{15}{*}{6} & \multirow{15}{*}{2-class} \\
  & global\_flow (visual) & & & & & \\
  & local\_flow (visual) & & & & & \\
  & face\_num (visual) & & & & & \\
  & volume (auditory) & & & & & \\
  & pitch (auditory) & & & & & \\
  & delta\_volume (auditory) & & & & & \\
  & speech (language) & & & & & \\
  & onset (language) & & & & & \\
  & gpt2\_surprisal (language) & & & & & \\
  & word\_length (language) & & & & & \\
  & word\_gap (language) & & & & & \\
  & word\_index (language) & & & & & \\
  & word\_head\_pos (language) & & & & & \\
  & word\_part\_speech (language) & & & & & \\
\midrule
 \textbf{Undefined(ECoG or SEEG)} &  Seizure Detection & MAYO & 5000 $\to$ 500Hz & Var.(16--72)& 8 & 2-class \\

\bottomrule
\end{tabular}
\end{table}

\subsection{Pretraining Setup and Model Scaling}\label{appendix_subsec:pretraining_setup} 
Training experiments were conducted across two high-performance computing configurations. The primary server consisted of nodes each equipped with a single 2.8 GHz AMD EPYC Milan 7543P 32-core CPU and four NVIDIA A100 GPUs, which was more heavily utilized throughout the training process. For large model variants, we additionally employed a secondary server equipped with dual Intel Xeon Platinum 8480+ processors (112 cores total) and eight NVIDIA H200 GPUs with 144GB memory each. Training experiments were conducted using either 16,32,128 A100 GPUs or 32, 24, 16 H200 GPUs depending on the experimental configuration. We maintained a fixed global batch size of 192 across all training runs, with the per-GPU batch size adjusted dynamically based on the number of nodes employed.

We \textcolor{black}{varied the model size by modifying} the hidden dimension of the transformer, resulting in sizes of 13M, 51M, 115M, 203M, 813M, 1.83B parameters, while keeping the depth fixed at 12 layers. This \textcolor{black}{capacity adjustment} leverages the benefits of $\mu$ parameterization for stable training across different model sizes. DIVER-1 was implemented on the Python 3.12.3 and Pytorch 2.6.0 + cuda version 12.4.
To enhance training efficiency, we employed DeepSpeed ZeRO Stage 2, BF16 precision. Optimization was performed using a custom implementation of the DeepSpeed's MuAdam optimizer \citep{muparamyang2022tensor} with utilizing DeepSpeed's FusedAdam backend \citep{rasley2020deepspeed} for computational efficiency and learning rate calibration. A cosine annealing learning rate scheduler with warm-up restarts was applied, with cycle length matching the total training steps and minimum learning rate set to 0.01× the initial rate.

\subsection{Spatio-Temporal Resampling}
\label{appendix_subsec:str}

STR is applied during self-supervised pretraining before masking and reconstruction. Each pretraining example is a 30\,s iEEG segment sampled at 500\,Hz. For DIVER-1-1s, the patch length is \(P=500\) samples, yielding \(N_0=30\) temporal patches; for DIVER-1-0.1s, the patch length is \(P=50\) samples, yielding \(N_0=300\) temporal patches.

Given a segment with \(C\) contacts, STR first defines the maximum sampled channel and temporal-token budgets as
\[
C_{\max}=\min(C,32), \qquad N_{\max}=\min(N_0,30).
\]
We then independently sample
\[
u_C,u_N \sim \mathrm{Beta}(3,1),
\]
and set
\[
C'=\max(1,\lceil C_{\max}u_C\rceil), \qquad
N'=\max(1,\lceil N_{\max}u_N\rceil).
\]
The \(\mathrm{Beta}(3,1)\) distribution has density concentrated toward larger values, so the model usually observes large subviews while still being exposed to smaller channel subsets and shorter temporal contexts. We sample \(C'\) contacts uniformly without replacement. For the temporal axis, we sample a contiguous crop of \(N'\) patches by drawing a start index
\[
\tau \sim \mathrm{Uniform}\{1,\ldots,N_0-N'+1\},
\]
and retaining patches \(\{\tau,\ldots,\tau+N'-1\}\). The resulting subview has shape \(C'\times N'\).

This resampling serves two purposes. First, random contact subsets expose the encoder to the variable electrode coverage typical of clinical iEEG, where patients differ in channel count, anatomical sampling, and implant geometry. Second, random temporal crops expose the encoder to variable context lengths. The cap \(N'\le 30\) also bounds the maximum number of electrode--time tokens to \(32\times 30\), keeping sequence-length-dependent attention compute comparable across DIVER-1-1s and DIVER-1-0.1s. In particular, without this cap, DIVER-1-0.1s would process up to 300 temporal patches per 30\,s segment, increasing attention cost substantially relative to DIVER-1-1s.

After STR, masked reconstruction is performed only on the retained \(C'\times N'\) token grid. STR is used only during pretraining; during downstream finetuning and evaluation, each task uses its native temporal window and available channel set.

\subsection{DIVER Architecture and Pretext Task Hyperparameter Setting}\label{appendix_subsec:diver_arch_pretext}

\textcolor{black}{\textbf{Architecture setting} Table \ref{tab:diver_hyperparams} lists the detailed architectural hyperparameter settings used for DIVER-1 pretraining.}

\textbf{FFT, STFT setting}
For the FFT, we used a window size of 500 time points with a sampling frequency of 500 Hz. A cutoff frequency of 200 Hz was applied, and the FFT amplitudes were converted to absolute values, normalized, and then compressed using a log(1 + x) transform. For the STFT, we employed a \textcolor{black}{multi-resolution} approach with window sizes of 200 and 100 time points, respectively. Each window was shifted with 50\% overlap and tapered with a Hann window function. Consistent with the FFT settings, a cutoff frequency of 200 Hz was applied, and the STFT amplitudes were converted to absolute values, normalized, and compressed using the log(1 + x) transform.

\textbf{MDRO setting} The total objective is a weighted reconstruction loss over masked patches and domains, \(\mathcal{L}_{\text{total}} = \sum_{(c,n)\in\mathcal{M}}\sum_{k\in\mathcal{K}}\lambda_k \mathcal{L}_{\text{MSE}}(\mathbf{y}^{(k)}_{c,n}, \hat{\mathbf{y}}^{(k)}_{c,n})\), where \(\mathcal{K}=\{\mathrm{raw},\mathrm{FFT},\mathrm{STFT}\}\), \(\lambda_k\) denotes the domain weight, and \(\mathcal{M}\) denotes masked patch indices. We use \((1,0.1,1)\) for DIVER-1-1s and \((1,1,0)\) for DIVER-1-0.1s, choosing weights to keep losses on comparable numerical scales; for \(P=50\), the STFT term is omitted because the window is too short for a meaningful spectrogram target.

\begin{table}[h]
\centering
\small
\caption{\textcolor{black}{\textbf{Hyperparameters for DIVER-1 pretraining.} Two model variants were trained for inputs with 1s and 0.1s patch size respectively. The 1s and 0.1s models share all settings except for patch size, patchwise CNN embedding settings, and SSL weights. Some hyperparameters are defined as a function of $d_{\text{model}}$, which we vary across \{256, 512, 768, 1024, 2048, 3072\}.}}
\label{tab:diver_hyperparams}
\begin{tabular}{lcc}
    \toprule
    & Hyperparameters & Settings \\
    \midrule
    
    \multirow{4}{*}{Input \& Masking} 
        & Patch size & 500 (1s) \\
        &            & 50 (0.1s) \\
        & Mask ratio & 0.5 \\
        & Masking type & Patch random \\
    \midrule
    
    \multirow{9}{*}{\shortstack{Patch Encoder\\(CNN)}} 
        & Intermediate channel & $d_{\text{model}}/8$ (1s) \\
        & ($C_{\text{inter}}$) & $d_{\text{model}}/16$ (0.1s) \\
        & Input dimension & \{1, $C_{\text{inter}}$, $C_{\text{inter}}$\} \\
        & Output dimension &  \{$C_{\text{inter}}$, $C_{\text{inter}}$, $C_{\text{inter}}$\} \\
        & Stride & \{64, 3, 3\} (1s) \\
        &        & \{4, 3, 3\} (0.1s) \\
        & Kernel size & \{63, 3, 3\} \\
        & Padding & \{31, 1, 1\} \\
        & Depth & 3 \\
    \midrule
    
    \multirow{2}{*}{\shortstack{Patch Encoder\\(Spectral)}} 
        & Spectral FFT size & $d_{\text{model}}/2 + 1$ \\
        & Spectral dropout & 0.1 \\
    \midrule
    
    \multirow{5}{*}{\shortstack{STCPE}} 
        & STCPE dimension & $d_{\text{model}}/8$ \\
        & STCPE layers & 1 \\
        & STCPE heads & $d_{\text{model}}/256$ \\
        & STCPE $d_{ff}$ & $d_{\text{model}}/2$ \\
        & Time window size & 7 \\
    \midrule

    \multirow{4}{*}{\shortstack{Positional\\Embedding}} 
        & Channel type dimension & $d_{\text{model}}/4$ \\
        & Embedding style & CPE (Learnable) \\
        & Temperature & 2000 \\
        & Scale & 1/256 \\
    \midrule
    
    \multirow{7}{*}{Transformer} 
        & Model dimension & $d_{\text{model}}$ \\
        & Layers & 12 \\
        & Heads & $d_{\text{model}}/32$ \\
        & Feed-forward dimension & $4*d_{\text{model}}$ \\
        & Activation & SiLU \\
        & Attention type & Flash attention \\
        & Dropout & 0.1 \\
    \midrule

    \multirow{4}{*}{SSL Head} 
        & Domain & Time, FFT, STFT \\
        & Loss weight ($\lambda_{\text{Time}}$) & 1.0 \\
        & Loss weight ($\lambda_{\text{FFT}}$)  & 0.1 (1s) / 1.0 (0.1s) \\
        & Loss weight ($\lambda_{\text{STFT}}$) & 1.0 (1s) / 0.0 (0.1s) \\
    \midrule
    
    \multirow{1}{*}{Training} 
        & Parameterization & $\mu$P \\
        
    \bottomrule
\end{tabular}
\end{table}

\FloatBarrier
\subsection{Pretraining Hyperparameter Search using $\mu$-Parameterization ($\mu$P) and $\mu$Transfer}\label{appendix_subsec:pretraining_hyperparam}

We employed a two-stage hyperparameter optimization approach to determine optimal learning rate (lr) and weight decay (wd) values; grid search followed by optuna optimization. The search was conducted using a 50M parameter model, a 12-layer architecture with 512-dimensional attention layers. Hyperparameter searches were performed over 2 epochs to balance computational efficiency with reliable performance estimation.\\
\textbf{Stage1: Grid search} We evaluated 30 distinct lr/wd combinataions across four stages of increasing specificity (Table ~\ref{tab:gridsearch}), identifying lr=6.0e-03 and wd=1.0e-06 as the best configuration for \textsc{$\textsc{DIVER-1-1s}_{\text{ Small}}$}. This configuration was also used as the starting point for the Stage 2 search of \textsc{$\textsc{DIVER-1-0.1s}_{\text{ Small}}$}.

\begin{table}
\footnotesize
\centering
\caption{\textbf{Stage 1: Grid search hyperparameter exploration.}}
\label{tab:gridsearch}
\begin{tabular}{l|l}
\hline
\textbf{Search Step} & \textbf{Hyperparameter Values} \\ \hline
Initial Learning Rate Exploration 
& Learning rate: \{1e-5, 1e-4, 1e-3, 1e-2\} (wd=1e-2) \\ \hline

Weight Decay Exploration 
& Weight decay: \{1e-7, 1e-6, 1e-5, 1e-4, 1e-3, 1e-2, 1e-1\} (lr=1e-3) \\ \hline

Refined Learning Rate Search 
& Learning rate: \{2e-4, 3e-4, 5e-4, 8e-4, 2e-3, 3e-3, 5e-3, 6e-3, 8e-3\} \\ \hline

Cross-combinations 
& lr=6e-3 with wd: \{1e-6, 1e-5, 1e-4, 1e-3, 1e-1\} \\ \hline
\end{tabular}
\end{table}

\textbf{Stage2: Optuna Optimization} We further refined the hyperparameters using Optuna\citep{optuna} or bayesian hyperparameter optimization. The search space was defined as $\pm 1$ order of magnitude around the best grid search configurations (range: $\times 0.1$ to $\times 10$), with 50 trials conducted to systematically explore this refined hyperparameter space. The optimal hyperparameter settings identified through Optuna optimization are presented in Table~\ref{tab:optimal_hyperparameters}.

\begin{table}[h]
\centering
\caption{\textbf{Optimal learning rate and weight decay}}
\label{tab:optimal_hyperparameters}
\begin{tabular}{lcccc}
\hline
\textbf{Models} & \textbf{Granularity} & \textbf{Learning Rate} & \textbf{Weight Decay} \\
\hline
$\textsc{DIVER-1-1s}_{\text{ Small}}$ & 1s & 2.30e-03 & 2.17e-07  \\
$\textsc{DIVER-1-0.1s}_{\text{ Small}}$ & 0.1s & 4.91e-03 & 3.75e-06 \\
\hline
\end{tabular}
\end{table}

\subsection{$\mu$-parameterization ($\mu P$)}\label{appendix_subsec:mu_param}
\begin{figure}[h]
\begin{center}
\includegraphics[width=1\linewidth]{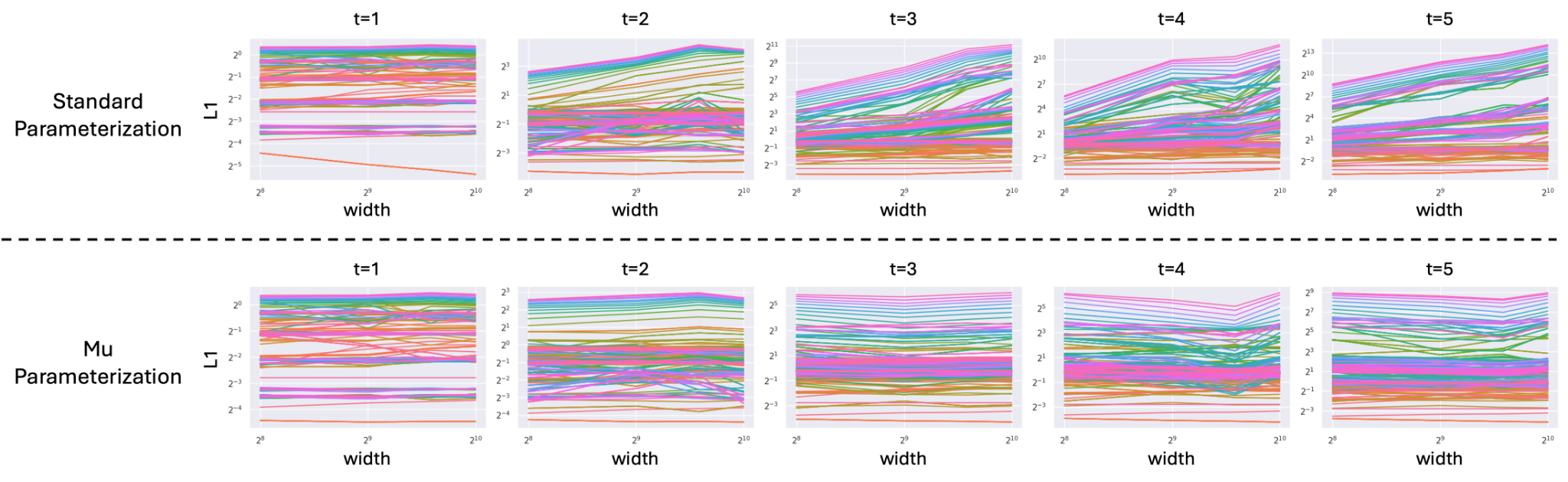}
\end{center}
\caption{
    \textbf{Verification of the $\mu P$ implementation.}
    The L1 norm of activation vectors (y-axis) is plotted against model width (x-axis) for five training timesteps (\textit{t}=1 to \textit{t}=5) across four different widths ($256, 512, 768, 1024$).
    \textbf{(Top Row)} With standard parameterization, activation norms are unstable and diverge as model width increases.
    \textbf{(Bottom Row)} In contrast, our $\mu$P implementation yields stable activation norms that are independent of model width. This confirms the model is correctly parameterized, a critical prerequisite for successful hyperparameter transfer via $\mu$Transfer.
}
\label{fig:mup_verification}
\end{figure}

The stable initialization shown in the \textbf{Figure \ref{fig:mup_verification}} translates directly to stable training dynamics. We confirmed this by tracking the training loss while \textcolor{black}{varying the model width} across the same four \textcolor{black}{configurations}. With $\mu P$ enabled, the training loss remained low and stable for all model sizes. In stark contrast, training without $\mu P$ led to severe instability; the loss diverged rapidly as the model grew, with values exploding to over 100 for the largest width. This empirical result demonstrates that our use of $\mu P$ was essential for reliably training larger models in our scaling experiments.


\subsection{Finetuning Setup}\label{appendix finetuning setup}\label{appendix_subsec:finetuning_setup} 
\textcolor{black}{For Neuroprobe,} DIVER-1 was finetuned at lr = 2e-3 and wd = 1e-2, batch size 32, with AdamW \textcolor{black}{for both frozen and full-finetuning. For the MAYO task, frozen models were fine-tuned using the same learning rate and weight decay as above, whereas full fine-tuning employed separate hyperparameters for each model size (lr = 1.19e-3, wd = 6.18e-1 for 256; lr = 4.19e-3, wd = 1.45e-2 for 512; lr = 7.14e-4, wd = 2.84e-1 for 768; lr = 1.40e-4, wd = 3.44e-2 for 1024; lr = 8.65e-4, wd = 9.90e-2 for 2048; selected via a learning-rate and weight-decay search based on the validation set of subject 1, fold 1).} 

Neuroprobe \citep{zahorodnii2025neuroprobe} reports the leaderboard performance for PopT and BrainBERT. The non-foundation-model baselines (CNN, MLP, and linear models) were computed from STFT spectrograms after Laplacian rereferencing and are also reported on the leaderboard. For Barista, we performed the evaluation using the original codebase, with the default settings: a learning rate of 1e-3 for the classifier and 1e-4 for the encoder, a batch size of 128, and the AdamW optimizer. However, while the original paper pretrained the model using 3-second windows, for the Neuroprobe evaluation we forced the input window to 1 second for evaluation.
 The MAYO seizure dataset is an extended dataset that we constructed to match Brant’s minimum input length, and therefore required training on the baseline models. BrainBERT was set at lr = 1e-3 for the classifier with AdamW, as in the original paper \citep{wang2023brainbert}, with batch size 32. The features in time [l-5:l+5] were concatenated along the channel dimension. 
\textcolor{black}{Brant was set at lr = 1e-4 for the classifier and 1e-7 for encoder layers, with betas=(0.9, 0.999), eps=1e-8, batch size 4 and Adam, same as their publicly released code.} 
\textcolor{black}{We could not evaluate PopT on MAYO because LPI coordinates are not available. Brant was pretrained with a 6 s patch and a total of 15 patches (90 s), so we were therefore unable to evaluate it on the neuroprobe (1 s), and there is a mismatch with the original pretrained context in MAYO (6 s).}

For all models other than BaRISTA, we trained for 40 epochs using a CosineAnnealing scheduler, applied the same validation split to each training set, and used early stopping when the validation AUROC did not improve for 10 epochs.
BaRISTA was trained for 30 epochs using the original codebase, with their default setting as learning-rate schedule consisting of a 5-epoch linear warmup followed by exponential decay.


\section{Scaling Law}\label{Appendix:Scaling Law}
\subsection{Scaling Law Experiment Details}\label{appendix_subsec:scaling_law_exp}

Pretraining loss curves for the trained models are presented in Figure \ref{fig:losscurve} for the DIVER-1-1s model family and Figure \ref{fig:losscurve_p50} for the DIVER-1-0.1s model family. We trained separate instances for each epoch to obtain the corresponding loss curves.   Importantly, we employed cosine annealing with warmup, where the decay schedule depends on the total number of training epochs. Therefore, extracting the loss at epoch 2 from a model trained for 32 epochs differs from the loss at epoch 2 of a model trained for only 2 epochs, as the learning rate trajectories diverge under these configurations.

\begin{figure}[htbp]
    \centering
    \includegraphics[width=1.0\textwidth]{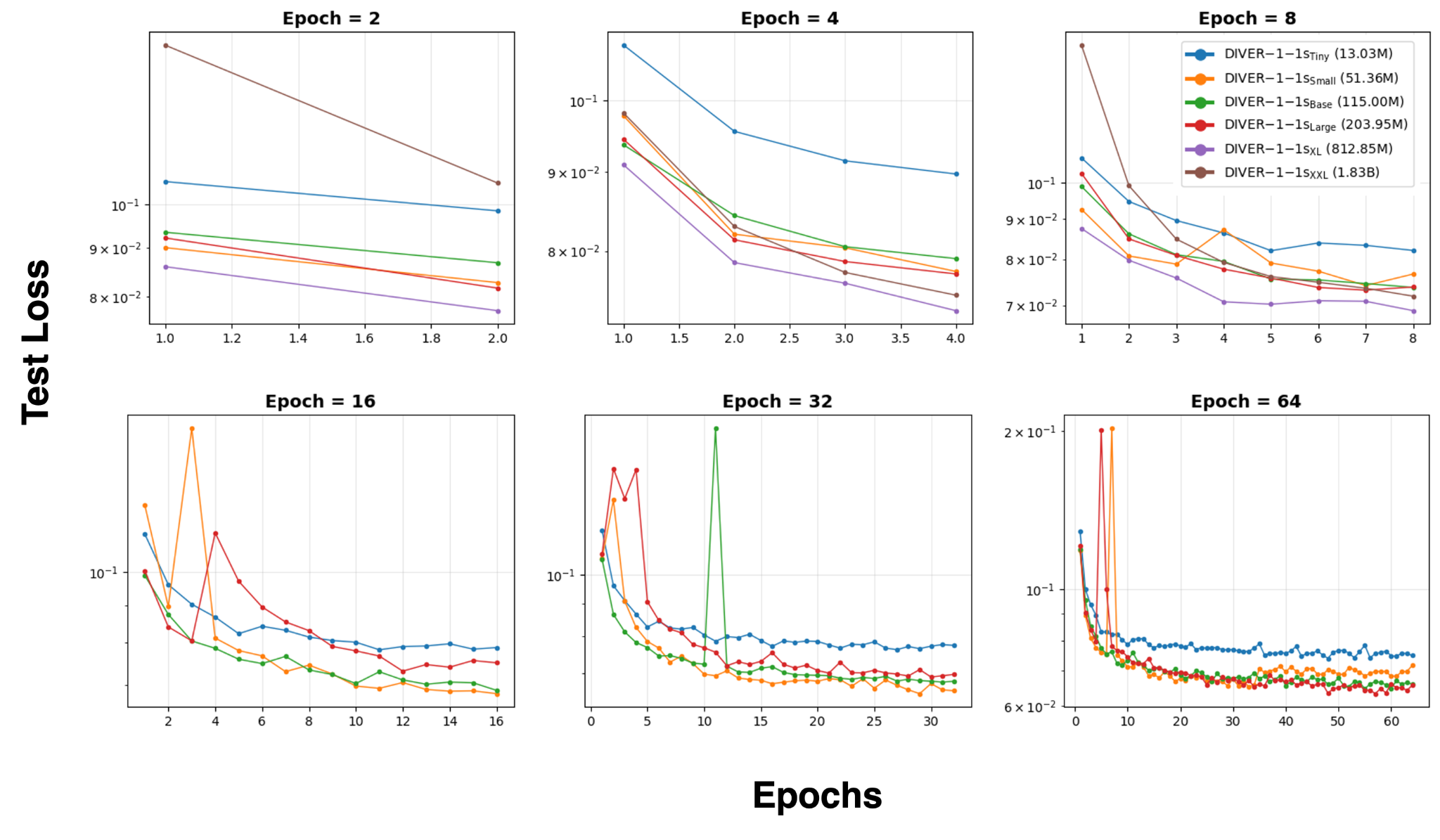}
    \caption{Loss curve\textcolor{black}{s} of \textcolor{black}{the DIVER-1-1s} model family. \textcolor{black}{Test loss across epochs is shown.}}
    \label{fig:losscurve}
\end{figure}

\begin{figure}[htbp]
    \centering
    \includegraphics[width=1.0\textwidth]{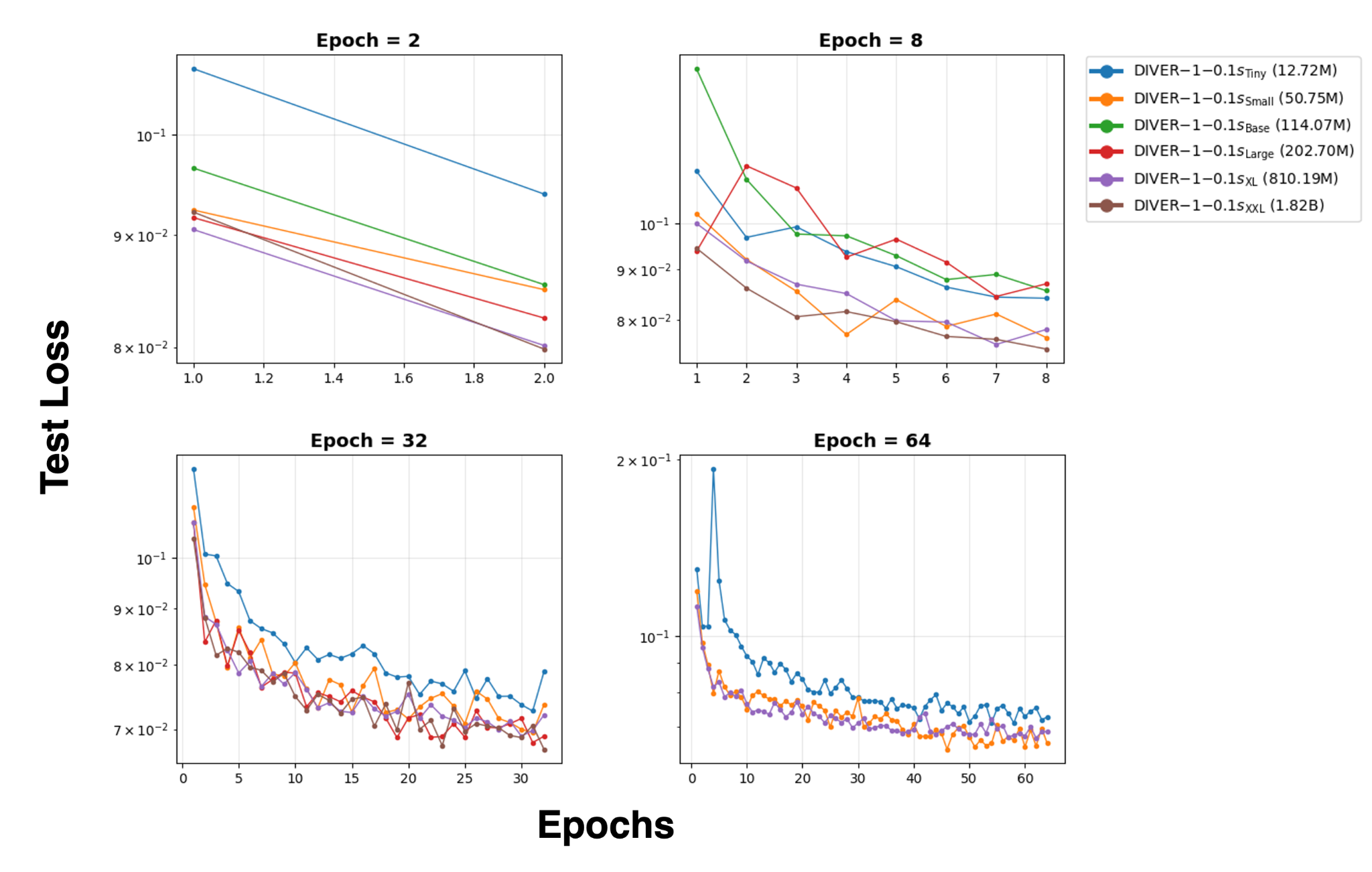}
    \caption{\textcolor{black}{Loss curve\textcolor{black}{s} of \textcolor{black}{the DIVER-1-0.1s} model family. Test loss across epochs is shown.}}
    \label{fig:losscurve_p50}
\end{figure}

\textcolor{black}{\textbf{Unique number of tokens}} First of all, the estimated total number of unique tokens is $U_D = \text{number of data sample} \times \text{number of tokens per sample}$. The number of tokens per sample can be expressed as $\text{number of channels} \times \text{number of timestamps} \times \text{proportion of unmasked patches}$.

The total number of data samples is 636,480. Since we randomly sampled from a 32 channel x 30 timestamp token grid using Beta(3,1) distribution for both axis, we estimate the number of tokens per sample as $32 \times 30 \times 0.75 \times 0.75=540$ tokens. Thus, we get $U_D=636,480 \times 540$ for the data.

\textbf{Compute}
To investigate compute scaling properties, we conducted systematic pretraining experiments across different epoch counts for the 1-second granularity models. For shorter training regimes (1, 2, 4, and 8 epochs), we pretrained all six model variants: $\textsc{DIVER-1-1s}_{\text{ Tiny}}$, $\textsc{DIVER-1-1s}_{\text{ Small}}$, $\textsc{DIVER-1-1s}_{\text{ Base}}$, $\textsc{DIVER-1-1s}_{\text{ Large}}$, $\textsc{DIVER-1-1s}_{\text{ XL}}$, and $\textsc{DIVER-1-1s}_{\text{ XXL}}$. For longer training regimes (16, 32, and 64 epochs), computational constraints limited our experiments to the four smaller model variants: $\textsc{DIVER-1-1s}_{\text{ Tiny}}$, $\textsc{DIVER-1-1s}_{\text{ Small}}$, $\textsc{DIVER-1-1s}_{\text{ Base}}$, and $\textsc{DIVER-1-1s}_{\text{ Large}}$.

\textcolor{black}{
For the DIVER-1-0.1s model family with 0.1-second granularity models, we followed a similar training protocol. We trained six model variant for epochs 2, 8 and five model variants for epochs 32, excluding $\textsc{DIVER-1-0.1s}_{\text{ Base}}$ and three model variants for epochs 64, excluding $\textsc{DIVER-1-0.1s}_{\text{ Base}}$, $\textsc{DIVER-1-0.1s}_{\text{ Large}}$ and $\textsc{DIVER-1-0.1s}_{\text{ XXL}}$.}

\textbf{Data Size} Data size scaling was done in 1, 3, 9, 24, 50, 90, 100\% of data on $\textsc{DIVER-1-1s}_{\text{ Small}}$ for 2 epochs with hyperparameters fixed as $lr=6.0e-03, wd=1.0e-06$.$\text{Number of token} = \text{number of data sample} \times \text{number of tokens per sample}$, while the total number of data samples was 636,480.
Number of tokens per sample estimated as 540, given that we randomly sampled from a 32×30 token grid using Beta(3,1) distribution. \textcolor{black}{A detailed explanation can be found in the aforementioned \textbf{Unique number of tokens} section.}

\textbf{Model Size} We fixed the number of epochs to 2. The models varied by their width, number of layers, and patch size. The detailed experiment conditions are on table~\ref{tab:model_params_pflops}. We observed that the models with different number of layers or patch size show different scaling behavior, so we fitted them separately.

\textbf{Number of Subjects}
Subject scaling experiments were done with datasets containing 2, 4, 5, 8, 10, 15, 16 subjects respectively, while maintaining a constant dataset size.$\textsc{DIVER-1-0.1s}_{\text{ Small}}$ trained for 2 epochs, due to compute constraint.

\textbf{Data-constrained Scaling Law} We trained a total of 31 DIVER-1-1s models with varying parameter counts and numbers of training epochs, while keeping the dataset fixed. The hidden dimension was fixed to 12 layers. Since different granularity led to different scaling behavior, we experimented on two granularity conditions and fitted them separately. 


\subsection{Extended Kaplan~\citep{kaplan2020scaling} Scaling Law Results}\label{appendix_subsec:extended_kaplan_ieeg}

We tested models at the 1-second \textcolor{black}{and 0.1-second} granularity. For 1-second granularity models, all six models listed in Table \ref{tab:model_params_pflops} as DIVER-1-1s were tested for epochs 2, 4, 8. At epochs 16, 32, and 64, evaluation was conducted on the following four models: $\textsc{DIVER-1-1s}_{\text{ Tiny}}$, $\textsc{DIVER-1-1s}_{\text{ Small}}$, $\textsc{DIVER-1-1s}_{\text{ Base}}$, and $\textsc{DIVER-1-1s}_{\text{ Large}}$. At earlier epochs (2, 4, and 8), the general trend showed decreasing loss as model size increased. However, as observed in $\textsc{DIVER-1-1s}_{\text{ XXL}}$, larger models exhibited substantially higher loss when trained with only a few epochs. This confirms what was also suggested by the data-constrained scaling isoplots: training very large models with insufficient updates is ineffective. Another possibility is that the aspect ratio of $\textsc{DIVER-1-1s}_{\text{ XXL}}$ (256) places it outside the region where loss remains stable. Future work should therefore evaluate larger models within the aspect-ratio regime where stable loss behavior is maintained.

\subsection{Data-Constrained Scaling Law Results}\label{appendix_subsec:data_constrained_ieeg}

\begin{table}[h]
\centering
\footnotesize
\caption{\textcolor{black}{\textbf{Fitted data-constrained scaling law parameters for DIVER-1-1s and DIVER-1-0.1s model families.}}} 
\label{tab:fit_values}
\begin{tabular}{lcc}
\toprule
 & DIVER-1-1s & DIVER-1-0.1s \\
\midrule
$A$              & 19.217 & 101.52 \\
$B$              & 57.065 & 1.1550  \\
$E$              & 0.0092 & 0.0030 \\
$\alpha$         & 0.3773  & 0.5248  \\
$\beta$          & 0.3504  & 0.1246  \\
$R_D^*$          & 9.5372  & 19.705 \\
$R_N^*$          & 3.3850  & 0.7191  \\
\midrule
$R^2$ (linear)   & 0.7858  & 0.7575  \\
$R^2$ (log)      & 0.8152  & 0.7718  \\
\bottomrule
\end{tabular}
\end{table}

\begin{figure}[htbp]
    \centering
    \includegraphics[width=\textwidth]{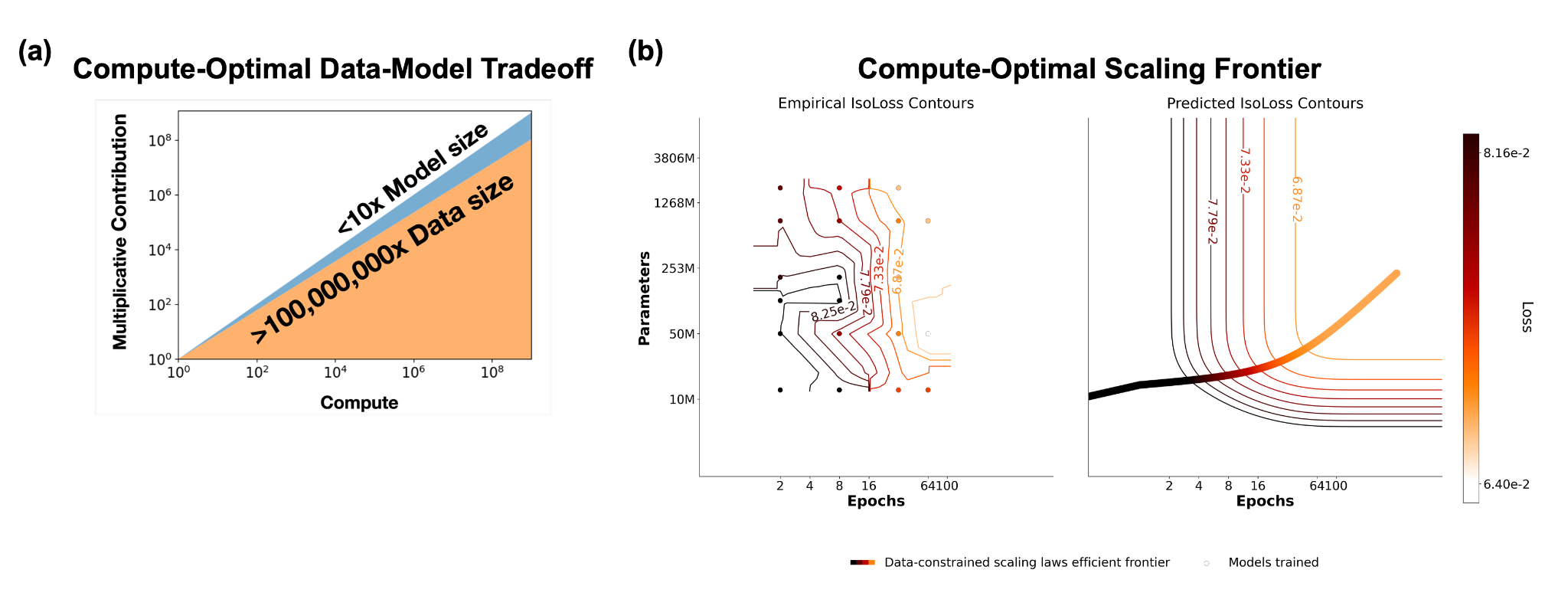}
    \caption{
    \textbf{Compute-optimal scaling of DIVER-1-0.1s. }(a) Optimal data–model allocation when training duration is fixed. With training fixed to two passes, we use the fitted DIVER-1-0.1s scaling law to estimate how the compute-optimal dataset size and parameter count should grow as total compute increases. (b) IsoLoss contours for the DIVER-1-0.1s model family. (Left) Twenty models with 0.1s patches were trained across varying epochs and parameter counts; iso-loss contours are obtained by linear interpolation between measured data points. (Right) Corresponding contours predicted by the fitted scaling law. The fading line denotes the minimum-loss configuration for each compute budget.
    }
    \label{fig:isoloss_patch50}
\end{figure}

Table~\ref{tab:fit_values} reports the fitted data-constrained scaling-law parameters for both DIVER-1-1s and DIVER-1-0.1s. The exponents $\alpha$ and $\beta$ govern the marginal benefit of scaling parameters and data, respectively.  For DIVER-1-1s, our values ($\alpha = 0.377$, $\beta = 0.350$) are similar in magnitude and comparable to language-domain results ($\alpha = 0.348$, $\beta = 0.366$;~\citep{besiroglu2024chinchilla}).  
The Kaplan-stye allocation in Figure~\ref{fig:ComputeOptimal}(a) is computed at fixed two-epochs training where data is effectively fresh ($R_D \approx 1$); under this fresh-data regime the data component naturally exerts greater influence on the loss, whereas the data-constrained fit aggregates across multi-epoch runs and yields balanced exponents.
DIVER-1-0.1s shows a more parameter-leaning exponent profile in isolation ($\alpha = 0.525$, $\beta = 0.125$), but its dataset-axis fit is less stable under our sampling density ($R^{2} \approx 0.77$); the small $\beta$ should therefore be read as reduced fit reliability rather than evidence that data scaling is unhelpful at 0.1s.

Importantly, the characteristic half-lives $R_D^\ast = 9.5$ and $R_N^\ast = 3.3$ quantify diminishing returns under repeated data and excessive parameters. The relatively larger $R_D^\ast$ implies that repeated data remains useful for many epochs before saturation, whereas the smaller $R_N^\ast$ suggests that the benefit of adding parameters decays more quickly. Together, these results suggest that gains are most effectively pursued by scaling dataset repetition while keeping model sizes small.


\textcolor{black}{
Figure \ref{fig:ComputeOptimal}(a) compares fixed-duration compute-optimal data--model allocation for  DIVER-1-0.1s models against the Kaplan language-model fit~\cite{kaplan_scaling_2020}. Although iEEG and text tokens are not directly comparable, the relative allocation differs strikingly across domains. Both DIVER-1-1s(Figure~\ref{fig:Scaling}(a) and DIVER-1-0.1s(Figure~\ref{fig:isoloss_patch50}(a)) exhibit a strongly data-leaning regime, in which optimal scaling devotes the overwhelming majority of additional compute to dataset size ($>5\times10^{7}\times$ for 1s model and $>10^{8}\times$ for 0.1s model) while model size grows by less than two orders of magnitude ($<20\times$ and $<10\times$, respectively). In contrast, the Kaplan LLM fit is strongly parameter-dominated, allocating roughly $3.7\times10^{6}\times$ to model size against only ${\sim}270\times$ to data. The two iEEG patch granularities therefore behave qualitatively similarly to each other and qualitatively differently from the language-model regime, suggesting that compute-optimal scaling for iEEG signals favors data scaling over parameter scaling---consistent with the limited per-sample information density of neural recordings relative to text.}

\textcolor{black}{The empirical isoLoss contours in Figure \ref{fig:ComputeOptimal} show less smoothness compared to the original scaling law paper\citep{data_constrained_scaling_muennighoff2023scaling}, primarily due to sampling density. While the original study used ~93 model configurations with dense sampling across all loss ranges, we evaluated ~30 configurations. Despite this visual difference, our empirical isoLoss contours (Figure \ref{fig:ComputeOptimal}(b) and Figure~\ref{fig:isoloss_patch50}(b), left panel) align well with the predicted contours (Figure \ref{fig:ComputeOptimal}(b) and Figure~\ref{fig:isoloss_patch50}(b), right panel), demonstrating that data-constrained scaling laws generalize to neural data.} Although our scaling fits are estimated on DIVER-1, the setup is representative of current EEG/iEEG FMs: recent reviews report that 82\% use Transformer backbones \cite{liu2026eegfoundationmodels} and 70\% use generation-based SSL\cite{shen2026brain4fms}. Other objectives, tokenizations, and corpora may change the exact exponents, but the intrinsic constraints of iEEG suggest that the qualitative recipe should persist: scale data first, train longer second, and scale parameters last.

\begin{figure}[htbp]
    \centering
    \includegraphics[width=1.0\textwidth]{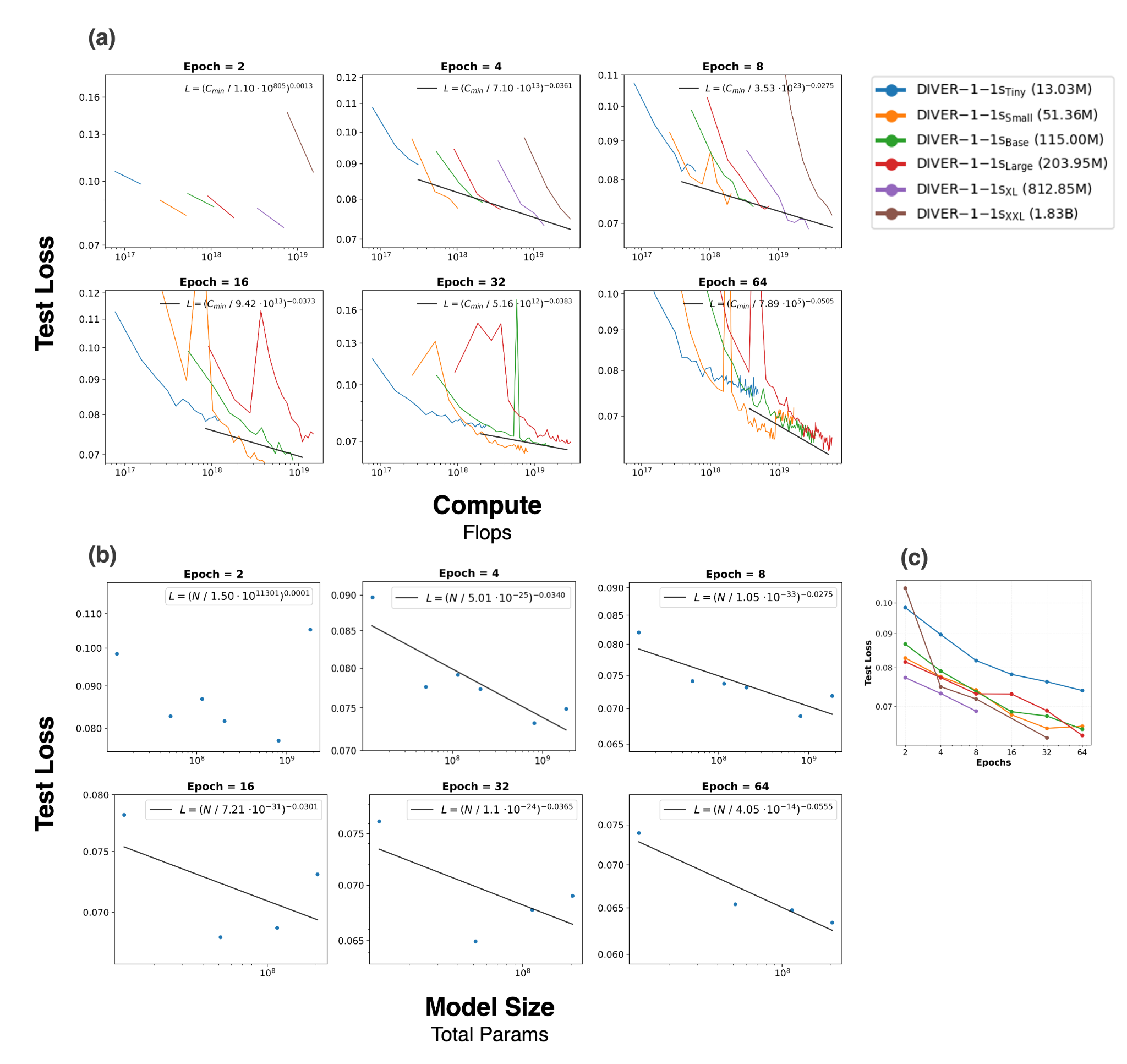}
    \caption{\textcolor{black}{Scaling law extended results for the DIVER-1-1s family. Compute scaling and model size scaling plots are given for models trained for 2, 4, 8, 16, 32, and 64 epochs.(a) Compute scaling and (b) model size scaling plots are given for models trained for 2, 8, 32, and 64 epochs. (c) Epoch scaling plot of the models reported in Fig.~\ref{fig:Scaling} (d). The same training runs are reused, with losses re-plotted against parameters and dataset size in log-log scale.}}
    \label{fig:scaling_extended_result}
\end{figure}

\begin{figure}[htbp]
    \centering
    \includegraphics[width=1.0\textwidth]{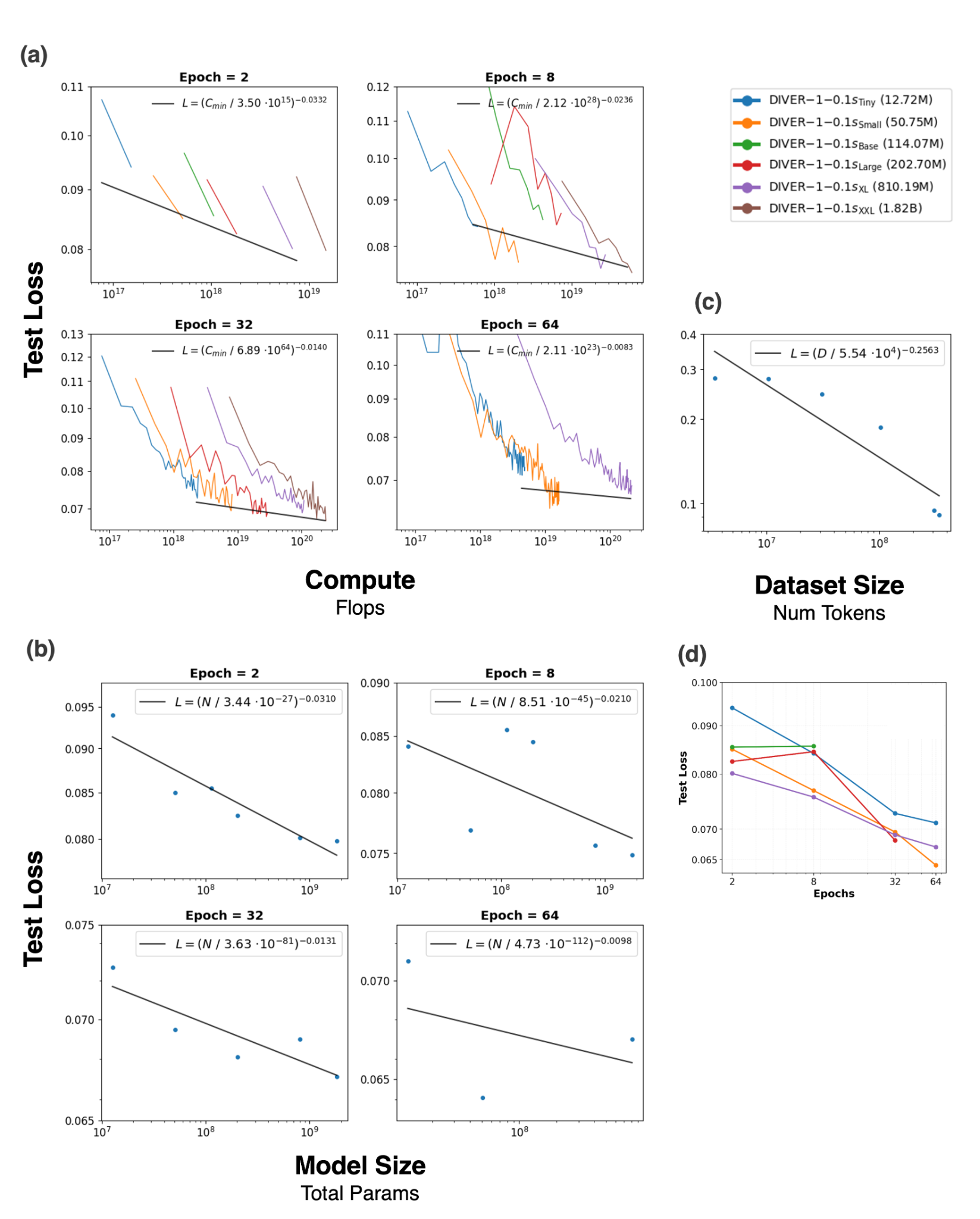}
    \caption{\textcolor{black}{Scaling law extended results for the DIVER-1-0.1s family. (a) Compute scaling and (b) model size scaling plots are given for models trained for 2, 8, 32, and 64 epochs. (c) Dataset size scaling plots are given for models trained for 2 epochs. (d) Epoch scaling plot of the models reported in Fig.~\ref{fig:isoloss_patch50}. The same training runs are reused, with losses re-plotted against parameters and dataset size in log-log scale.}}
    \label{fig:scaling_extended_result_p50}
\end{figure}

\section{Extended Results}\label{appendix_sec:extended_results}
\subsection{Comprehensive DIVER Downstream Task Results}\label{appendix_subsec:comprehensive_diver_downstream}

\textcolor{black}{We present the detailed numerical values. The performance table is provided in Table~\ref{table:ieeg_performance_total} and Table~\ref{table:neuroprobe_full_results}.}
\textcolor{black}{
For Neuroprobe binary-label, we evaluated DIVER-1-0.1s (with $d_{\text{model}}=256$ and pretrained for 32 epochs) and DIVER-1-1s (with $d_{\text{model}}=256$ and pretrained for 32 epochs) in both frozen (linear probing) and fine-tuned  configurations against Linear Laplacian STFT, MLP Laplacian STFT, CNN Laplacian STFT, BrainBERT, PopT and BaRISTA baselines on Figure~\ref{fig:SOTAperf}. Table~\ref{table:neuroprobe_full_results} presents comprehensive results comparing DIVER with baseline models across all 15 Neuroprobe tasks.}
\textcolor{black}{Results are reported as mean AUROC $\pm$ SEM across subjects, trials, and cross-validation folds. DIVER consistently outperformed baseline models across the majority of tasks in both evaluation settings. The model demonstrated particularly strong performance on language-related tasks (speech decoding, word prediction, onset detection) and auditory tasks (volume, pitch), with finetuning providing additional gains over frozen features. These results validate that self-supervised pretraining on iEEG data produces representations that transfer effectively to diverse downstream neural decoding tasks.}
\textcolor{black}{For MAYO, we evaluated DIVER-1-1s (with $d_{\text{model}} = 256$, pretrained for 32 epochs) in both frozen (linear probing) and full-finetuning settings and DIVER-1-0.1s only in frozen setting due to GPU memory constraint. We could not evaluate PopT because the dataset does not contain any coordinates. By contrast, although our model is trained with 3D positional embeddings, it can handle missing position information by replacing the positional embedding with a zero vector for electrodes with unknown location. Among the different baselines (BrainBERT frozen, Brant frozen, and Brant full-finetuning), DIVER-1-1s achieved the best performance. Brant exhibited a substantial performance drop under full finetuning, likely because we deviated from its original pretraining configuration by using only a single patch, which can impair optimization when updating all parameters with original context windows (90s), whereas the frozen setting simply uses the fixed embedding.}

\begin{table}[h]
\centering
\color{black}
\caption{\textbf{Comparison of the DIVER-1 model with other baseline models.} We evaluated DIVER-1 models that were pretrained for 32 epochs on 100\% of the pretraining dataset. For Neuroprobe (1 s), we compare DIVER-1 with other baselines using an overall score defined as the mean AUROC averaged over all 15 tasks, subjects, trials, and folds. For MAYO (6 s), we compare DIVER-1 with Brant and BrainBERT; scores are reported as mean AUROC $\pm$ SEM across 8 subjects and 3 folds.}
\label{table:ieeg_performance_total}
\footnotesize
\begin{tabular}{lcc}
\toprule
 & Neuroprobe & Mayo \\
\midrule
linear (stft-laplacian)      & $0.660 \pm 0.007$ & - \\
MLP on stft-laplacian        & $0.656 \pm 0.007$ & - \\
CNN on stft-laplacian        & $0.669 \pm 0.007$ & - \\
PopT                         & $0.569 \pm 0.005$ & - \\
BaRISTA                      & $0.586 \pm 0.006$ & - \\
BrainBERT (frozen)           & $0.626 \pm 0.007$ & $0.748 \pm 0.038$ \\
Brant                        & -                 & $0.551 \pm 0.023$ \\
Brant (frozen)               & -                 & $0.757 \pm 0.042$ \\
\midrule
DIVER-1-0.1s                 & $0.665 \pm 0.008$ & - \\
DIVER-1-0.1s (frozen)        & $\mathbf{0.678 \pm 0.007}$ & $0.955 \pm 0.010$ \\
DIVER-1-1s                   & $0.655 \pm 0.007$ & $\mathbf{0.961 \pm 0.011}$ \\
DIVER-1-1s (frozen)          & $0.662 \pm 0.007$ & $0.935 \pm 0.012$ \\
\bottomrule
\end{tabular}
\end{table}

\begin{table}[t]
\centering
\caption{\textcolor{black}{\textbf{Downstream performance of each task in Neuroprobe.} We compare DIVER with existing models on comprehensive iEEG downstream tasks. We evaluated both the fine-tuned and frozen configurations of DIVER-1-0.1s and 1s (pretrained on iEEG for 32 epochs) against Linear Laplacian STFT, MLP Laplacian STFT, CNN Laplacian STFT, PopT, BaRISTA, BrainBERT, and Brant. Results are reported as mean AUROC $\pm$ SEM across subjects, trials, and folds. Overall, DIVER consistently outperformed baselines on 10 out of 15 tasks.}}
\label{table:neuroprobe_full_results}
\scriptsize
\color{black}
\begin{tabular}{lcccc}
\toprule
\textbf{Models} & \textbf{Overall} & \textbf{Sentence Onset} & \textbf{Speech} & \textbf{Volume} \\
\midrule
        Linear Laplacian STFT & $0.660 \pm 0.007$ & $0.888 \pm 0.015$ & $0.882 \pm 0.014$ & $0.717 \pm 0.028$ \\
        MLP Laplacian STFT & $0.656 \pm 0.007$ & $0.884 \pm 0.018$ & $0.877 \pm 0.018$ & $0.717 \pm 0.027$ \\
        CNN Laplacian STFT & $0.669 \pm 0.007$ & $0.886 \pm 0.015$ & $0.888 \pm 0.017$ & $\mathbf{0.757 \pm 0.028}$ \\
        PopT & $0.569 \pm 0.005$ & $0.767 \pm 0.025$ & $0.686 \pm 0.028$ & $0.569 \pm 0.012$ \\
        BaRISTA & $0.586 \pm 0.006$ & $0.739 \pm 0.031$ & $0.680 \pm 0.031$ & $0.560 \pm 0.016$ \\
        BrainBERT {\scriptsize(frozen)} & $0.626 \pm 0.007$ & $0.883 \pm 0.015$ & $0.887 \pm 0.015$ & $0.723 \pm 0.022$ \\
\midrule
        DIVER-1-0.1s & $0.665 \pm 0.008$ & $0.924 \pm 0.009$ & $0.897 \pm 0.013$ & $0.688 \pm 0.020$ \\
        DIVER-1-0.1s {\scriptsize(frozen)} & $\mathbf{0.678 \pm 0.007}$ & $\mathbf{0.930 \pm 0.009}$ & $\mathbf{0.897 \pm 0.013}$ & $0.719 \pm 0.021$ \\
        DIVER-1-1s & $0.655 \pm 0.007$ & $0.890 \pm 0.013$ & $0.848 \pm 0.015$ & $0.634 \pm 0.017$ \\
        DIVER-1-1s {\scriptsize(frozen)} & $0.662 \pm 0.007$ & $0.912 \pm 0.010$ & $0.885 \pm 0.013$ & $0.700 \pm 0.019$ \\
\bottomrule
\end{tabular}
\vspace{5pt}
\begin{tabular}{lcccc}
\toprule
\textbf{Models} & \textbf{Delta Volume} & \textbf{Voice Pitch} & \textbf{Word Position} & \textbf{Inter-word Gap} \\
\midrule
        Linear Laplacian STFT & $0.756 \pm 0.021$ & $0.584 \pm 0.014$ & $0.740 \pm 0.022$ & $0.613 \pm 0.013$ \\
        MLP Laplacian STFT & $0.747 \pm 0.021$ & $0.585 \pm 0.013$ & $0.728 \pm 0.021$ & $0.597 \pm 0.013$ \\
        CNN Laplacian STFT & $0.762 \pm 0.019$ & $\mathbf{0.618 \pm 0.015}$ & $0.734 \pm 0.021$ & $0.583 \pm 0.014$ \\
        PopT & $0.649 \pm 0.018$ & $0.514 \pm 0.005$ & $0.610 \pm 0.016$ & $0.533 \pm 0.008$ \\
        BaRISTA & $0.725 \pm 0.024$ & $0.513 \pm 0.009$ & $0.691 \pm 0.025$ & $0.559 \pm 0.011$ \\
        BrainBERT {\scriptsize(frozen)} & $0.677 \pm 0.019$ & $0.564 \pm 0.013$ & $0.620 \pm 0.013$ & $0.526 \pm 0.008$ \\
\midrule
        DIVER-1-0.1s & $\mathbf{0.814 \pm 0.020}$ & $0.561 \pm 0.007$ & $\mathbf{0.799 \pm 0.017}$ & $0.634 \pm 0.014$ \\
        DIVER-1-0.1s {\scriptsize(frozen)} & $0.803 \pm 0.019$ & $0.592 \pm 0.008$ & $0.798 \pm 0.017$ & $0.636 \pm 0.011$ \\
        DIVER-1-1s & $0.808 \pm 0.020$ & $0.554 \pm 0.009$ & $0.796 \pm 0.016$ & $\mathbf{0.642 \pm 0.013}$ \\
        DIVER-1-1s {\scriptsize(frozen)} & $0.775 \pm 0.020$ & $0.576 \pm 0.008$ & $0.762 \pm 0.018$ & $0.616 \pm 0.009$ \\
\bottomrule
\end{tabular}
\vspace{5pt}
\begin{tabular}{lcccc}
\toprule
\textbf{Models} & \textbf{GPT-2 Surprisal} & \textbf{Head Word Pos} & \textbf{Part of Speech} & \textbf{Word Length} \\
\midrule
        Linear Laplacian STFT & $0.608 \pm 0.013$ & $0.603 \pm 0.009$ & $0.601 \pm 0.012$ & $0.617 \pm 0.013$ \\
        MLP Laplacian STFT & $0.604 \pm 0.014$ & $0.602 \pm 0.011$ & $0.597 \pm 0.012$ & $0.615 \pm 0.014$ \\
        CNN Laplacian STFT & $0.597 \pm 0.013$ & $0.614 \pm 0.010$ & $0.621 \pm 0.018$ & $0.636 \pm 0.014$ \\
        PopT & $0.561 \pm 0.010$ & $0.540 \pm 0.005$ & $0.525 \pm 0.006$ & $0.533 \pm 0.006$ \\
        BaRISTA & $0.566 \pm 0.009$ & $0.566 \pm 0.011$ & $0.562 \pm 0.011$ & $0.578 \pm 0.012$ \\
        BrainBERT {\scriptsize(frozen)} & $0.557 \pm 0.014$ & $0.567 \pm 0.009$ & $0.576 \pm 0.012$ & $0.586 \pm 0.010$ \\
\midrule
        DIVER-1-0.1s & $0.618 \pm 0.009$ & $0.617 \pm 0.009$ & $0.607 \pm 0.013$ & $0.638 \pm 0.013$ \\
        DIVER-1-0.1s {\scriptsize(frozen)} & $\mathbf{0.629 \pm 0.011}$ & $\mathbf{0.623 \pm 0.010}$ & $0.627 \pm 0.013$ & $\mathbf{0.646 \pm 0.014}$ \\
        DIVER-1-1s & $0.612 \pm 0.009$ & $0.609 \pm 0.010$ & $\mathbf{0.631 \pm 0.013}$ & $0.637 \pm 0.016$ \\
        DIVER-1-1s {\scriptsize(frozen)} & $0.609 \pm 0.011$ & $0.609 \pm 0.009$ & $0.617 \pm 0.011$ & $0.625 \pm 0.013$ \\
\bottomrule
\end{tabular}
\vspace{5pt}
\begin{tabular}{lcccc}
\toprule
\textbf{Models} & \textbf{Global Flow} & \textbf{Local Flow} & \textbf{Frame Brightness} & \textbf{Num of Faces} \\
\midrule
        Linear Laplacian STFT & $0.630 \pm 0.015$ & $0.615 \pm 0.014$ & $0.528 \pm 0.024$ & $0.519 \pm 0.013$ \\
        MLP Laplacian STFT & $0.629 \pm 0.018$ & $0.617 \pm 0.015$ & $0.524 \pm 0.021$ & $0.521 \pm 0.014$ \\
        CNN Laplacian STFT & $\mathbf{0.638 \pm 0.017}$ & $\mathbf{0.630 \pm 0.012}$ & $\mathbf{0.536 \pm 0.021}$ & $\mathbf{0.530 \pm 0.012}$ \\
        PopT & $0.541 \pm 0.009$ & $0.517 \pm 0.010$ & $0.498 \pm 0.012$ & $0.493 \pm 0.007$ \\
        BaRISTA & $0.518 \pm 0.011$ & $0.522 \pm 0.010$ & $0.518 \pm 0.013$ & $0.498 \pm 0.005$ \\
        BrainBERT {\scriptsize(frozen)} & $0.609 \pm 0.013$ & $0.599 \pm 0.013$ & $0.497 \pm 0.016$ & $0.516 \pm 0.011$ \\
\midrule
        DIVER-1-0.1s & $0.594 \pm 0.012$ & $0.584 \pm 0.012$ & $0.486 \pm 0.020$ & $0.518 \pm 0.008$ \\
        DIVER-1-0.1s {\scriptsize(frozen)} & $0.628 \pm 0.012$ & $0.616 \pm 0.012$ & $0.495 \pm 0.021$ & $0.525 \pm 0.010$ \\
        DIVER-1-1s & $0.592 \pm 0.009$ & $0.574 \pm 0.010$ & $0.490 \pm 0.011$ & $0.502 \pm 0.007$ \\
        DIVER-1-1s {\scriptsize(frozen)} & $0.614 \pm 0.015$ & $0.601 \pm 0.012$ & $0.502 \pm 0.023$ & $0.523 \pm 0.011$ \\
\bottomrule
\end{tabular}
\end{table}

\subsection{Performance Evaluation across DIVER-1 Model configurations}\label{appendix_subsec:perf_eval_across_diver1} 
We \textcolor{black}{first} evaluated DIVER-1 models' performances \textcolor{black}{in Neuroprobe,} based on different model configurations including the patch size, Laplacian re-referencing and training settings. The 0.1s model outperformed the 1s model (Table \ref{table:DIVERvariations_with_linearandfullft}) in \textcolor{black}{4 tasks in Neuroprobe}. Considering that the Neuroprobe dataset consists of 1-second samples and its tasks require classifying  short-timescale features such as speech and onset, the effectiveness of 0.1s model may be explained. \textcolor{black}{Additionally, the model without Laplacian re-referencing generally showed degraded performance, indicating the effectiveness of the pre-processing method . Further, we examined the effect of pretraining by comparing the performance of DIVER-1-0.1s with diverse training settings. The model trained from scratch showed significantly degraded performance than the full-finetuned and backbone-frozen models.} 
\textcolor{black}{Such results indicate the efficacy of pretraining.} Specifically, the model with a frozen backbone showed
the highest performance, except for the speech task.
Table\ref{table:DIVERvariations_with_linearandfullft} shows that  models’ downstream performance improve when Laplacian re-referencing is used. PopT and BrainBERT were pretrained on Laplacian re-referenced signals, and their downstream performances were also derived under that setting. Even though our model was trained on raw signals (not referenced), it was better with Laplacian referencing (Table \ref{table:DIVERvariations_with_linearandfullft}). Therefore, we use Laplacian re-referencing as the default setting for finetuning on iEEG downstream tasks.

\begin{table}[h]
\centering
\caption{\textcolor{black}{\textbf{Performance evaluation between various DIVER model configurations in Neuroprobe (iEEG tasks).} We first compare  $\textsc{DIVER-1}_{\text{ Tiny}}$  models by patch size, Laplacian re-referencing, and training settings. The model from scratch was trained on only four tasks (speech, onset, volume, pitch) due to computational constraints, consequently; model evaluation is limited to these four tasks. For the backbone-frozen models, 0.1s variants with different sizes were trained on four tasks, whereas the tiny model was trained on all Neuroprobe tasks. The results are reported as mean AUROC ± SEM across multiple subjects, trials, and folds. All models, except the one trained from scratch, were pretrained for 32 epochs on 100\% of the pretraining dataset.During the development of Neuroprobe, the train/validation/test split was updated, and these results were obtained using the previous split, prior to September 2025.} }
\label{table:DIVERvariations_with_linearandfullft}
\tiny
\begin{tabular}{lcccccc}
    & speech & onset & volume & pitch  \\ 
\midrule
$\textsc{DIVER-1-1s}_{\text{ Tiny}}$ & $ 0.828\pm0.016 $  & $0.885 \pm 0.012 $  & $ 0.634\pm0.018 $  & $ 0.551\pm0.009 $ \\
$\textsc{DIVER-1-0.1s}_{\text{ Tiny}}$  & $ \textbf{0.900} \pm \textbf{0.011} $  & $ 0.924 \pm 0.009 $  & $ 0.699\pm0.020 $  & $0.563 \pm 0.007 $  &  \\
$\textsc{DIVER-1-0.1s}_{\text{ Tiny}}$ {\scriptsize(w.o. laplacian)} & $ 0.862 \pm 0.018 $  & $ 0.901\pm 0.013$  &$ 0.662\pm0.018 $  &$ 0.533\pm 0.004 $  & \\
$\textsc{DIVER-1-0.1s}_{\text{ Tiny}}$ {\scriptsize(from scratch)} & $ 0.832\pm 0.014 $  &$ 0.872\pm 0.010 $  &$ 0.622\pm0.016 $  &$ 0.554\pm0.006 $  &  \\ 
\midrule
$\textsc{DIVER-1-0.1s}_{\text{ Tiny}}$ {\scriptsize(frozen)}  & $ 0.896\pm0.012 $  &$ \textbf{0.930}\pm\textbf{0.008} $  &$ \textbf{0.717}\pm\textbf{0.018} $  &$ \textbf{0.589}\pm\textbf{0.007} $  & \\ 
$\textsc{DIVER-1-0.1s}_{\text{ Small}}$ {\scriptsize(frozen)} & $ 0.888 \pm 0.012$  &$ 0.926\pm0.009 $  &$ 0.705\pm0.016 $  &$ 0.578\pm0.006 $  & \\
$\textsc{DIVER-1-0.1s}_{\text{ Large}}$ {\scriptsize(frozen)} & $ 0.890\pm0.0126 $  &$ 0.928\pm 0.009$  &$ 0.710\pm0.017 $  &$ 0.581\pm0.007 $  &  \\
$\textsc{DIVER-1-0.1s}_{\text{ XL}}$  {\scriptsize(frozen)} & $ 0.893\pm0.012 $  &$ 0.930\pm0.009 $  &$ 0.713\pm0.017 $  &$ 0.582\pm0.007 $  & \\ 
\bottomrule
\end{tabular}
\end{table}
\

\textcolor{black}{For MAYO (seizure detection), we compare within only the 1s model family, as MAYO uses a 6s window that is too long for the context length of the 0.1s models. Under the 0.1s setting, we were able to evaluate only the frozen Tiny model, while the remaining variants could not be run. We compared the frozen and full finetuning model in each model size for DIVER-1-1s in Table~\ref{table:mayo_dataset_scaling_and_frozen_vs_fullft}. In contrast to the Neuroprobe results, we found that full fine-tuning outperformed the frozen for the Tiny, Small and XL model, even though it was slightly worse for the Large and Base models. Since our preprocessing clips signal amplitudes above a certain threshold (200 \si{\micro\volt}), so pretrained-dataset's distribution can differ from seizure data, which contains spikes with much larger amplitudes; under this distribution shift, fully fine-tuned models may therefore tend to achieve better performance. For the linear baseline, the highest performance was obtained with the Large model, whereas under full-finetuning the Tiny model achieved the best performance. }

\begin{table}[h]
\centering
\color{black}
\caption{\textbf{Performance evaluation between various DIVER model configurations in MAYO } DIVER 1 s models were trained for each model size, and we compared the frozen and full fine-tuning variants. 
The results are reported as mean AUROC ± SEM across multiple subjects and folds. All models, except were pretrained for 32 epochs on 100\% of the pretraining dataset.}
\label{table:mayo_dataset_scaling_and_frozen_vs_fullft}
\footnotesize
\begin{tabular}{lcccccc}
    & Tiny & Small & Base & Large & XL & \\ 
\midrule
frozen & $ 0.935\pm0.012 $  & $0.904 \pm 0.019 $  & $ 0.911\pm0.017 $  & $ 0.937 \pm 0.015 $  &  $ 0.914\pm 0.018$  & \\
full-finetuned  & $ \mathbf{0.961 \pm 0.011} $  & $ \mathbf{0.927 \pm 0.020} $  & $ \mathbf{0.905 \pm0.026} $  & $\mathbf{0.934 \pm 0.014} $ & $ \mathbf{0.947 \pm 0.019} $  &  \\
\bottomrule
\end{tabular}
\end{table}

\subsection{Interpretation Results}\label{appendix_subsec:interpretation_results} 

Figure~\ref{fig:umap} shows a visualization of representation analysis on downstream datasets using UMAP~\citep{mcinnes2018umap}. We examine the embeddings obtained from the pretrained DIVER-1-0.1s model on the test set of speech, onset, and volume tasks in neuroprobe dataset without any finetuning. The results indicate that DIVER learned meaningful iEEG representations from pretraining, thereby capturing label-relevant structure in downstream datasets even in the absence of finetuning.

\begin{figure}[!htbp]
  \centering
  \includegraphics[width=\textwidth]{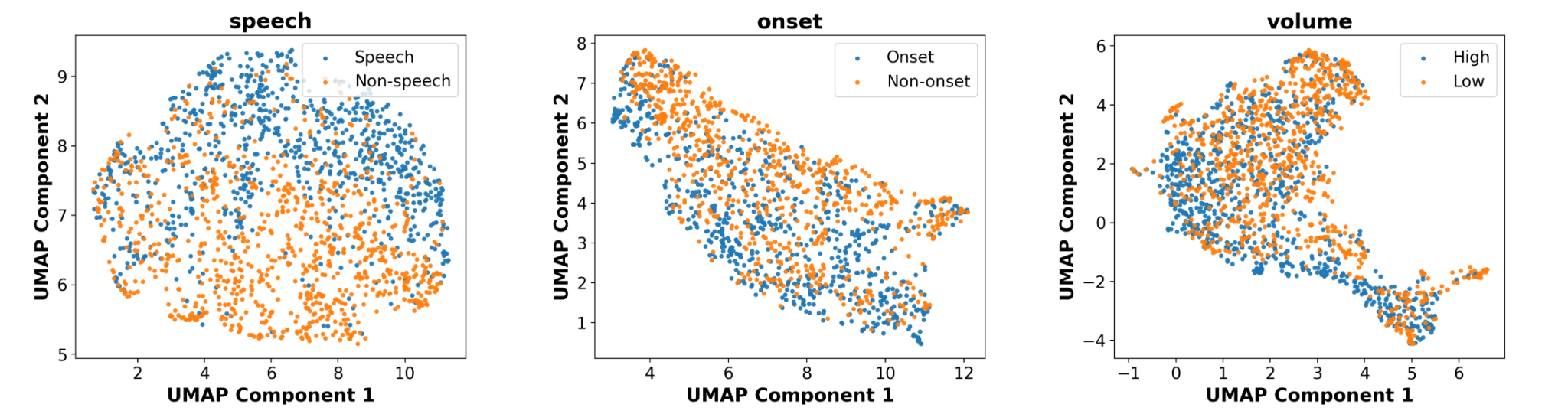}
  \caption{Visualizations of representations on neuroprobe downstream tasks. Each plot shows test set embeddings from one fold of a single trial from a single subject.}
  \label{fig:umap}
\end{figure}


To obtain an interpretable salience measure of the model, we applied attention rollout \citep{abnar2020quantifying}, which aggregates attention weights across all attention layers. Although DIVER’s any-variate attention produces a high-dimensional matrix, the attention rollout yielded a rank-1 matrix across all subjects and sessions. We therefore used a single representative row, reflecting a uniform attention distribution across keys independent of query. Figure~\ref{fig:brainmapping} shows task-dependent temporal and cortical attribution patterns derived from the attention rollout of DIVER-1-0.1s. 
Attention rollout weights were computed on test sets for four high AUROC scored Neuroprobe tasks (volume, pitch, GPT-2 surprisal, and word head position). We follow PopT’s method of mapping attention weights onto the cortical surface, normalizing within each subject and scaling by the test AUROC \citep{chau2025popt}. The resulting weights were averaged across subjects, trials, and folds. 

Across both auditory (volume, pitch) and language (GPT-2 surprisal, word head position) tasks, attention rollout exhibits two temporal peaks: an early window (0–300 ms) and a late window (800–1000 ms). The early peak coincides with word onset of each sample and engages core speech–language regions, including inferior temporal gyrus and Broca’s and Wernicke’s areas \citep{mesgarani2014phonetic}. The late peak shows task-specific spatial patterns. For auditory tasks, attribution remains concentrated in primary auditory cortex (Heschl’s gyrus and superior temporal gyrus). This is consistent with the possible continued auditory processing of subsequent words within the 1s window. In contrast, language tasks show increased engagement of temporal-frontal regions in the later window. This aligns with the known post-auditory latency of higher language networks \citep{hickok2007cortical, rauschecker2009maps}. 
Overall, these results indicate that DIVER attends to tokens in a temporally structured  manner with task-relevant spatial attribution. Notably, the right middle occipital gyrus remains consistently salient for GPT-2 surprisal, suggesting a contribution of visual-context representations to surprisal estimation \citep{friederici2011brain, dehaene2011unique}. Collectively, DIVER’s task-dependent attributions are consistent with established neural models of auditory and language processing \citep{hickok2007cortical}.

\begin{figure}[!b]
  \centering
  \includegraphics[width=\textwidth]{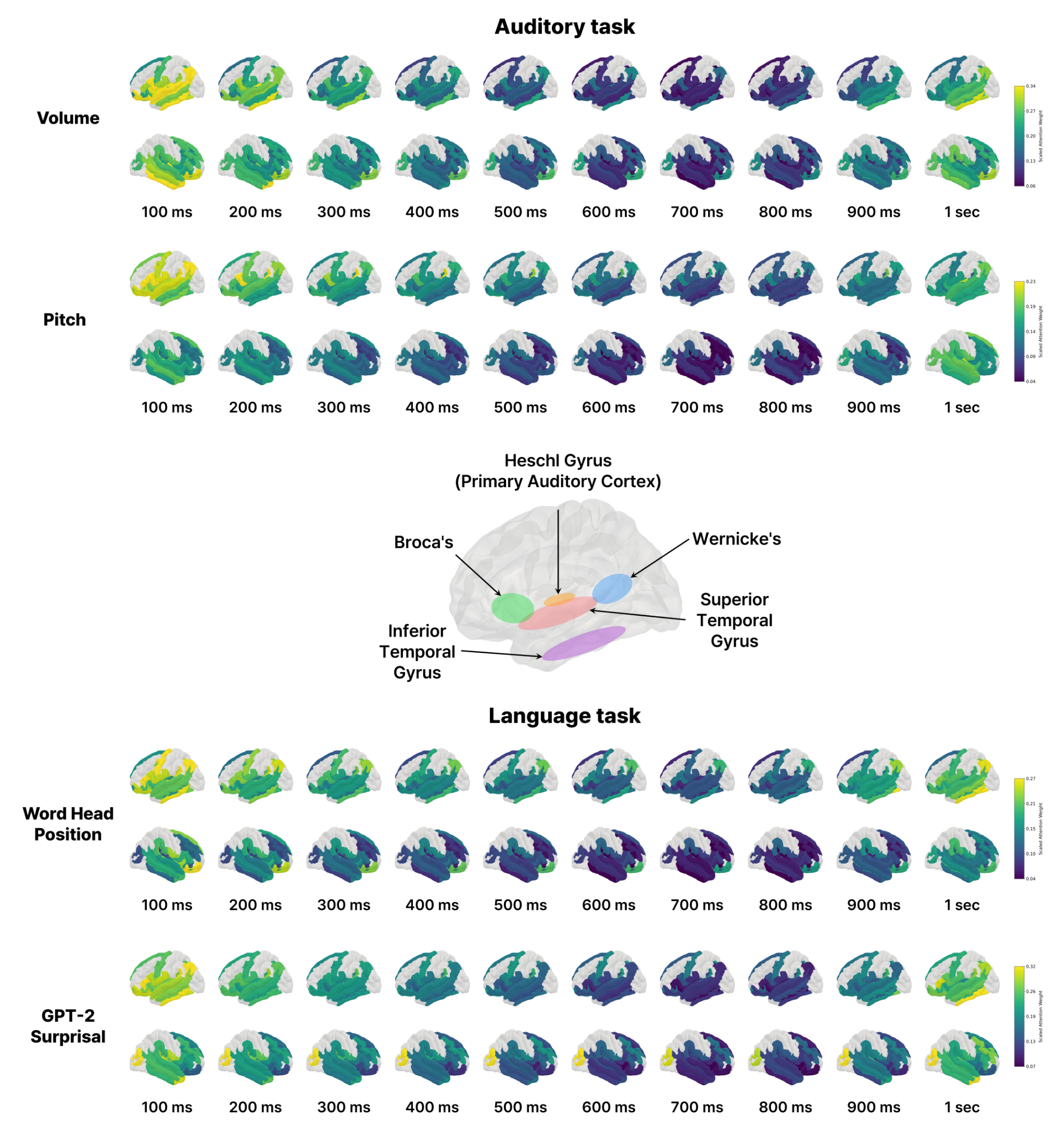}
  \caption{\textbf{Scaled attention rollout weights mapped onto the cortical surface for four Neuroprobe downstream tasks.} Each panel shows DIVER-1-0.1s scaled attention weights extracted via attention rollout and aggregated in 100~ms bins from 0~ms to 1~s relative to stimulus onset. For each task, the top row shows the left hemisphere and the bottom row shows the right hemisphere; brighter colors indicate larger scaled attention weights, which we interpret as greater model salience on the corresponding channel-temporal token for each task.}
  \label{fig:brainmapping}
\end{figure}


\section{Data Details}\label{appendix:data_details}
\subsection{Pretraining Dataset Description}\label{appendix_subsec:pretraining_dataset_description}
The following datasets were utilized for the pretraining of our DIVER models. The total pretraining time is 5,310 hours.

\begin{itemize}
    \item \textbf{AJILE12 (Annotated Joints in Long-term Electrocorticography)}~\citep{peterson2022ajile12}: An ECoG dataset from 12 epilepsy patients, recorded semi-continuously over 55 days. Signals were collected from $\geq$ 64 electrodes at 1 kHz sampling rate and paired with synchronized video-based 3D human pose estimation and annotated wrist-movement events.
    \item \textbf{Self-collected iEEG dataset}: An intracranial EEG dataset from 25 drug-resistant epilepsy patients  ($\sim$7 days, $\sim$168 h per subject) with long-term ECoG and sEEG recordings (mean 56.4~$\pm$~3.38 channels, sampled at 2 kHz) during naturalistic hospital behaviors.
\end{itemize}

\subsection{Finetuning Dataset Description}\label{appendix_subsec:finetuning_dataset_description}

\textcolor{black}{The following datasets were utilized for the downstream evaluation of our DIVER models. An overview of the dataset specifications and task definitions is provided in Table~\ref{table:overview_downstream}.}

\textbf{Neuroprobe}
Neuroprobe \citep{zahorodnii2025neuroprobe} is a large scale iEEG benchmarks with naturalistic labels during movie watching. 10 subjects watch 25 movies, age from 6 to 19. There are 3 types of evaluation in neuroprobe; single subject-single movie (WithinSession) (splits within the movies), single subject-different movie(CrossSession) (splits within subjects), different subject-different movie(CrossSubject). We evaluated the model in WithinSession. Additionally, Neuroprobe provides an option to subset subjects and trials. We used the LITE option (default configuration), which includes two movies per subject and a total of six subjects.  \footnote{PopT's performance on the same task differs between its original paper and its evaluation in the neuroprobe benchmark because neuroprobe implemented proper train/test splits across time so that no temporal leakage occurs between training and test sets, whereas the original PopT evaluation used random sampling that can lead to data contamination across temporal boundaries.} Detailed description of each task is provided below (adapted from \citep{zahorodnii2025neuroprobe}:

\begin{enumerate}
    \item \textbf{frame\_brightness} \textit{(visual)}: The mean brightness computed as the average HSV value over all pixels. Low (percentiles 0\%-25\%) vs High (75\%-100\%)
    \item \textbf{global\_flow} \textit{(visual)}: A camera motion proxy. The maximal average dense optical flow vector magnitude. Same as above.
    \item \textbf{local\_flow} \textit{(visual)}: A large displacement proxy. The maximal optical flow vector magnitude. Same as above.
    \item \textbf{face\_num} \textit{(visual)}: The maximum number of faces per frame during the word. 0, or $\geq$ 1.
    \item \textbf{volume} \textit{(auditory)}: Average root mean squared watts of the audio. Low (0\%-25\%) vs High (75\%-100\%).
    \item \textbf{pitch} \textit{(auditory)}: Average pitch of the audio. Same as above.
    \item \textbf{delta\_volume} \textit{(auditory)}: The difference in average RMS of the 500 ms windows pre- and post-word onset. Same as above.
    \item \textbf{speech} \textit{(language)}: Whether any speech is present in the given time interval.
    \item \textbf{onset} \textit{(language)}: Whether a new sentence starts in the interval, or there is no speech at all.
    \item \textbf{gpt2\_surprisal} \textit{(language)}: Negative-log transformed GPT-2 word probability (given preceding 20s of language context). Low (0\%-25\%) vs High (75\%-100\%).
    \item \textbf{word\_length} \textit{(language)}: Word length (ms). Same as above.
    \item \textbf{word\_gap} \textit{(language)}: Difference between previous word offset and current word onset (ms). Same as above.
    \item \textbf{word\_index} \textit{(language)}: The word index in its context sentence. The first word in the sentence (0), or other (1).
    \item \textbf{word\_head\_pos} \textit{(language)}: The relative position (left/right) of the word’s dependency tree head.
    \item \textbf{word\_part\_speech} \textit{(language)}: The word Universal Part-of-Speech (UPOS) tag. Verb (0), or other (1).
\end{enumerate}

\subsection{QAQC and preprocessing}
\label{appendix_subsec:QAQC_preproc}

All data underwent quality assessment and control (QAQC) and preprocessing with a philosophy of minimal intervention to retain as much original signal information as possible. For QAQC, we clipped amplitude values exceeding these normalization thresholds, only discarding electrodes when more than 3.33\% of samples required clipping and removing whole segments when more than 50\% of channels were compromised. \textcolor{black}{This conservative strategy enabled us to preserve substantially more usable data. Then we normalized signals by dividing by 200 \si{\micro\volt}. While \citet{jiang2024labram} (EEG model) applied normalization without QAQC and \citet{wang2024cbramod}  (EEG model) removed entire segments if even one timepoint exceeded 100 \si{\micro\volt}, we adopted a more conservative clipping approach to prevent data loss.

For preprocessing, we applied minimal filtering: a high-pass filter (0.5 Hz) to remove low-frequency drift, 60 - 120 - 180 notch filters for power line noise suppression, and no low-pass filtering to preserve high-frequency components. All datasets were resampled to 500 Hz and segmented into 30-second non-overlapping windows.}

\section{Comparison With Existing iEEG Foundation Models}\label{appendix:previous_EFM}

We compare DIVER-1 to all prior iEEG foundation models on the axes most relevant for scaling: parameter count, subject diversity, channels, dataset volume (channel-hours), and effective compute, the last reported as Scaled Epochs $=\Phi/352{,}035$ where $\Phi$ is total channel-hours processed during pretraining and $352{,}035$ is the DIVER-1 corpus size.

\begin{table}[h]
\centering
\caption{\textbf{Comparison with prior iEEG foundation models.}}
\label{tab:previous_EFM}
\begin{threeparttable}
\scriptsize
\renewcommand{\arraystretch}{1.3}
\setlength{\tabcolsep}{4pt}
\begin{tabularx}{\textwidth}{@{} l c c c >{\raggedright\arraybackslash}X c c c @{}}
\toprule
\textbf{Model} & \textbf{Modality} & \textbf{\makecell{Params}} & \textbf{\makecell{\#Subj.\\(sites)}} & \textbf{Cohort} & \textbf{\makecell{Volume\\(ch-hr)}} & \textbf{\makecell{Train.\\Epochs}} & \textbf{\makecell{Sc.\,Eps.\\on Ours\tnote{a}}}\\
\midrule
BrainBERT~\citep{wang2023brainbert} & sEEG & 43.18M & 10 (1) & Brain Treebank\tnote{b} & 4{,}551 & 39\tnote{*} & 0.51\\
Brant~\citep{zhang2023brant} & sEEG & 505.69M & 9 (1) & private clinical & 281{,}860\tnote{*} & 426\tnote{*} & 340.88\\
Du-IN~\citep{zheng2024discrete}\tnote{c} & sEEG & 4.38M & 12 (1) & private (Beijing)\tnote{c} & 150\tnote{c} & 400 & 0.17\\
BaRISTA~\citep{oganesian2025barista} & sEEG & 1M & 10 (1) & Brain Treebank\tnote{b} & 3{,}640\tnote{*} & 70 & 0.72\\
PopT~\citep{chau2025popt}\tnote{d} & sEEG & 20M\tnote{d} & 10 (1) & Brain Treebank\tnote{b} & 6{,}570\tnote{*} & n.r. & n.r.\\
\textbf{Ours} (DIVER-1) & ECoG+sEEG & 12.72M--1.83B & \textbf{37 (2)} & AJILE12\,+\,private (multi-site) & \textbf{352{,}035} & 64 & 64\\
\bottomrule
\end{tabularx}
\begin{tablenotes}
    \item[a] $\text{Sc.\,Eps.(Scaled Epochs)} = (\text{Volume}\!\times\!\text{Train.\,Epochs}) / 352{,}035$, normalising total channel-hours processed during pretraining by the DIVER-1 corpus size. This is a unit-consistent compute metric across models with different per-sample structures.
    \item[b] BrainBERT, BaRISTA, and PopT all pretrain on the same 10-subject sEEG corpus (Boston Children's Hospital; movie-watching). BrainBERT used the pre-release version; the corpus was subsequently released as ``Brain Treebank''. Volume differences reflect channel/session selection only.
    \item[c] Du-IN is a per-subject model; we report the SOTA configuration (62.70\% acc.) pretrained on full 15\,h recordings (12\,h non-task + 3\,h task). Multi-subject (``poms'', 59.18\%) and non-task-only (12\,h, 60.02\%) variants underperformed and are omitted.
    \item[d] PopT is trained on top of a frozen BrainBERT; the 20\,M count excludes the 43\,M BrainBERT base.
    \item[n.r.] Not reported. Specific information is missing from the original publication and cannot be accurately derived.
    \item[*] Estimated; derivations in Table~\ref{tab:estimates}.
\end{tablenotes}
\end{threeparttable}
\end{table}

\begin{table}[ht]
\centering
\scriptsize
\caption{\textbf{Estimation of model and training specifications.} We detail the assumptions and calculations used to derive data volume (channel-hours) and training epochs for prior models. Bold values represent the final estimates derived from the reported configurations.}
\label{tab:estimates}
\begin{threeparttable}
\begin{tabularx}{\textwidth}{l l c >{\raggedright\arraybackslash}X}
\toprule
\textbf{Model} & \textbf{Quantity} & \textbf{Est. Value} & \textbf{Justification / Derivation Method} \\
\midrule
\textbf{BrainBERT} & Train.\,Epochs & \textbf{39.06} & $U{=}500\text{k}, B{=}256, A{=}1$ \citep{wang2023brainbert}. Each sample is a 5\,s single-channel spectrogram ($h_s = 5/3600$\,ch-hr). \\
& & & $\Phi = U \cdot B \cdot A \cdot h_s = 1.78 \times 10^5$\,ch-hr. \\
& & & \textbf{Epochs} $= \Phi / 4{,}551 = \mathbf{39.06}$. \\
\midrule
\textbf{Brant} & Volume & \textbf{281,860} & Individual ch-hr summed across 9 subjects ($\sum c_i h_i$) using metadata from App.~E Table~7 (totaling 894 channels and 2,528 recording hours). This precise count is highly consistent with the reported ``$>10^{12}$ timestamps'' ($\ge 277,778$\,ch-hr) at 1,000\,Hz.\\
\addlinespace
& Train.\,Epochs & \textbf{425.74} & 
$U{=}750\text{k}, B{=}16, A{=}4 \Rightarrow 48 \times 10^6$ sample-passes. Based on 1,500 patches/sample ($L{=}15, C{\approx}100$) and $\tau{=}6$\,s, $h_s = 2.5$\,ch-hr. \\
& & & $\Phi = 48 \times 10^6 \cdot 2.5 = 1.20 \times 10^8$\,ch-hr. \\
& & & \textbf{Epochs} $= \Phi / 281{,}860 = \mathbf{425.74}$. \\
\midrule
\textbf{BaRISTA} & Volume & \textbf{3,640} & 35,089 segments of 3\,s totaling 29.2\,h \citep{oganesian2025barista}. Mean channel count $\bar{C} = 1{,}247 / 10 = 124.7$. \\
& & & \textbf{Volume} $= 29.2 \cdot 124.7 \approx \mathbf{3,640}$\,ch-hr. \\
\midrule
\textbf{PopT} & Volume & \textbf{6,570} & \textbf{Volume} $=$ 19 sessions $\times$ 2.07\,h/session $\times$ 167 mean ch/subj \citep{chau2025popt} $\approx \mathbf{6,570}$\,ch-hr. \\
\bottomrule
\end{tabularx}

\begin{tablenotes}[para,flushleft]
\scriptsize
\textit{Notation.} $U$: optimizer updates, $B$: minibatch size, $A$: gradient accumulation steps, $h_s$: per-sample channel-hours, $\Phi$: total ch-hr processed ($\Phi = U \cdot B \cdot A \cdot h_s$).
\textbf{Epochs:} Ratio of total throughput ($\Phi$) to dataset volume. \textbf{Scaled Epochs on DIVER$_\text{I}$:} $\Phi/352{,}035$, providing a unit-consistent comparison across heterogeneous sampling pipelines (e.g., single-channel spectrograms vs. multi-channel windows).
\end{tablenotes}
\end{threeparttable}
\end{table}


}{\typeout{Warning: appendix.tex not found; skipping appendix input.}}


\clearpage
\bibliographystyle{unsrtnat}
\bibliography{References}

@article{bommasani2021opportunities,
  title={On the opportunities and risks of foundation models},
  author={Bommasani, Rishi and Hudson, Drew A and others},
  journal={arXiv preprint arXiv:2108.07258},
  year={2021}
}

@article{besiroglu2024chinchilla,
  title={Chinchilla scaling: A replication attempt},
  author={Besiroglu, Tamay and Erdil, Ege and Barnett, Matthew and You, Josh},
  journal={arXiv preprint arXiv:2404.10102},
  year={2024}
}

@article{scangos2021closed,
  title={Closed-loop neuromodulation in an individual with treatment-resistant depression},
  author={Scangos, Katherine W and Khambhati, Ankit N and Daly, Patrick M and Makhoul, Ghassan S and Sugrue, Leo P and Zamanian, Hashem and Liu, Tony X and Rao, Vikram R and Sellers, Kristin K and Dawes, Heather E and others},
  journal={Nature medicine},
  volume={27},
  number={10},
  pages={1696--1700},
  year={2021},
  publisher={Nature Publishing Group US New York}
}

@article{khambhati2019functional,
  title={Functional control of electrophysiological network architecture using direct neurostimulation in humans},
  author={Khambhati, Ankit N and Kahn, Ari E and Costantini, Julia and Ezzyat, Youssef and Solomon, Ethan A and Gross, Robert E and Jobst, Barbara C and Sheth, Sameer A and Zaghloul, Kareem A and Worrell, Gregory and others},
  journal={Network Neuroscience},
  volume={3},
  number={3},
  pages={848--877},
  year={2019},
  publisher={MIT Press One Rogers Street, Cambridge, MA 02142-1209, USA journals-info~…}
}

@article{kaplan_scaling_2020,
    author  = {Kaplan, Jared and McCandlish, Sam and Henighan, Tom and Brown, Tom B. and Chess, Benjamin and Child, Rewon and Gray, Scott and Radford, Alec and Wu, Jeffrey and Amodei, Dario},
    title   = {Scaling laws for neural language models},
    journal = {arXiv preprint},
    year    = {2020},
    eprint  = {2001.08361},
    archivePrefix = {arXiv}
  }

@article{hoffmann_chinchilla_2022,
    author  = {Hoffmann, Jordan and Borgeaud, Sebastian and Mensch, Arthur and others},
    title   = {Training compute-optimal large language models},
    journal = {arXiv preprint},
    year    = {2022},
    eprint  = {2203.15556},
    archivePrefix = {arXiv}
  }

@article{su2024roformer,
  title={Roformer: Enhanced transformer with rotary position embedding},
  author={Su, Jianlin and Ahmed, Murtadha and Lu, Yu and Pan, Shengfeng and Bo, Wen and Liu, Yunfeng},
  journal={Neurocomputing},
  volume={568},
  pages={127063},
  year={2024},
  publisher={Elsevier}
}

@article{wang2024cbramod,
  title={CBraMod: A Criss-Cross Brain Foundation Model for EEG Decoding},
  author={Wang, Jiquan and Zhao, Sha and Luo, Zhiling and Zhou, Yangxuan and Jiang, Haiteng and Li, Shijian and Li, Tao and Pan, Gang},
  journal={arXiv preprint arXiv:2412.07236},
  year={2024}
}

@article{wang2024brain,
  title={Brain treebank: Large-scale intracranial recordings from naturalistic language stimuli},
  author={Wang, Christopher and Yaari, Adam and Singh, Aaditya and Subramaniam, Vighnesh and Rosenfarb, Dana and DeWitt, Jan and Misra, Pranav and Madsen, Joseph and Stone, Scellig and Kreiman, Gabriel and others},
  journal={Advances in Neural Information Processing Systems},
  volume={37},
  pages={96505--96540},
  year={2024}
}

@article{chen2024eegformer,
  title={Eegformer: Towards transferable and interpretable large-scale eeg foundation model},
  author={Chen, Yuqi and Ren, Kan and Song, Kaitao and Wang, Yansen and Wang, Yifan and Li, Dongsheng and Qiu, Lili},
  journal={arXiv preprint arXiv:2401.10278},
  year={2024}
}

@article{chau2025popt,
  title={Population Transformer: Learning population-level representations of neural activity},
  author={Chau, Geeling and Wang, Christopher and Talukder, Sabera and Subramaniam, Vighnesh and Soedarmadji, Saraswati and Yue, Yisong and Katz, Boris and Barbu, Andrei},
  journal={ArXiv},
  pages={arXiv--2406},
  year={2025}
}

@article{zhang2023brant,
  title={Brant: Foundation model for intracranial neural signal},
  author={Zhang, Daoze and Yuan, Zhizhang and Yang, Yang and Chen, Junru and Wang, Jingjing and Li, Yafeng},
  journal={Advances in Neural Information Processing Systems},
  volume={36},
  pages={26304--26321},
  year={2023}
}

@article{jiang2024labram,
  title={Large brain model for learning generic representations with tremendous EEG data in BCI},
  author={Jiang, Wei-Bang and Zhao, Li-Ming and Lu, Bao-Liang},
  journal={arXiv preprint arXiv:2405.18765},
  year={2024}
}

@article{woo2024moirai,
  title={Unified training of universal time series forecasting transformers},
  author={Woo, Gerald and Liu, Chenghao and Kumar, Akshat and Xiong, Caiming and Savarese, Silvio and Sahoo, Doyen},
  journal={International Conference on Machine Learning},
  year={2024},
  publisher={PMLR}
}

@article{stam2007phase,
  title={Phase lag index: assessment of functional connectivity from multi channel EEG and MEG with diminished bias from common sources},
  author={Stam, Cornelis J and Nolte, Guido and Daffertshofer, Andreas},
  journal={Human brain mapping},
  volume={28},
  number={11},
  pages={1178--1193},
  year={2007},
  publisher={Wiley Online Library}
}

@article{fries2005mechanism,
  title={A mechanism for cognitive dynamics: neuronal communication through neuronal coherence},
  author={Fries, Pascal},
  journal={Trends in cognitive sciences},
  volume={9},
  number={10},
  pages={474--480},
  year={2005},
  publisher={Elsevier}
}

@article{deco2011emerging,
  title={Emerging concepts for the dynamical organization of resting-state activity in the brain},
  author={Deco, Gustavo and Jirsa, Viktor K and McIntosh, Anthony R},
  journal={Nature reviews neuroscience},
  volume={12},
  number={1},
  pages={43--56},
  year={2011},
  publisher={Nature Publishing Group UK London}
}

@article{varela2001brainweb,
  title={The brainweb: phase synchronization and large-scale integration},
  author={Varela, Francisco and Lachaux, Jean-Philippe and Rodriguez, Eugenio and Martinerie, Jacques},
  journal={Nature reviews neuroscience},
  volume={2},
  number={4},
  pages={229--239},
  year={2001},
  publisher={Nature Publishing Group UK London}
}

@misc{bbrinkm2014upenn,
  title={UPenn and Mayo Clinic’s Seizure Detection Challenge},
  author={Bbrinkm, S and Cukierski, W},
  year={2014},
  publisher={Kaggle.[Online]. Available: https://kaggle. com/competitions/seizure-detection}
}

@article{kaplan2020scaling,
  title={Scaling laws for neural language models},
  author={Kaplan, Jared and McCandlish, Sam and Henighan, Tom and Brown, Tom B and Chess, Benjamin and Child, Rewon and Gray, Scott and Radford, Alec and Wu, Jeffrey and Amodei, Dario},
  journal={arXiv preprint arXiv:2001.08361},
  year={2020}
}

@misc{hoffmann2022chinchilla,
      title={Training Compute-Optimal Large Language Models}, 
      author={Jordan Hoffmann and Sebastian Borgeaud and Arthur Mensch and Elena Buchatskaya and Trevor Cai and Eliza Rutherford and Diego de Las Casas and Lisa Anne Hendricks and Johannes Welbl and Aidan Clark and Tom Hennigan and Eric Noland and Katie Millican and George van den Driessche and Bogdan Damoc and Aurelia Guy and Simon Osindero and Karen Simonyan and Erich Elsen and Jack W. Rae and Oriol Vinyals and Laurent Sifre},
      year={2022},
      eprint={2203.15556},
      archivePrefix={arXiv},
      primaryClass={cs.CL},
      url={https://arxiv.org/abs/2203.15556}, 
}

@article{data_constrained_scaling_muennighoff2023scaling,
  title={Scaling data-constrained language models},
  author={Muennighoff, Niklas and Rush, Alexander and Barak, Boaz and Le Scao, Teven and Tazi, Nouamane and Piktus, Aleksandra and Pyysalo, Sampo and Wolf, Thomas and Raffel, Colin A},
  journal={Advances in Neural Information Processing Systems},
  volume={36},
  pages={50358--50376},
  year={2023}
}

@article{muparamyang2022tensor,
  title={Tensor programs v: Tuning large neural networks via zero-shot hyperparameter transfer},
  author={Yang, Greg and Hu, Edward J and Babuschkin, Igor and Sidor, Szymon and Liu, Xiaodong and Farhi, David and Ryder, Nick and Pachocki, Jakub and Chen, Weizhu and Gao, Jianfeng},
  journal={arXiv preprint arXiv:2203.03466},
  year={2022}
}

@article{wang2023brainbert,
  title={BrainBERT: Self-supervised representation learning for intracranial recordings},
  author={Wang, Christopher and Subramaniam, Vighnesh and Yaari, Adam Uri and Kreiman, Gabriel and Katz, Boris and Cases, Ignacio and Barbu, Andrei},
  journal={arXiv preprint arXiv:2302.14367},
  year={2023}
}

@article{peterson2022ajile12,
  title={AJILE12: Long-term naturalistic human intracranial neural recordings and pose},
  author={Peterson, Steven M and Singh, Satpreet H and Dichter, Benjamin and Scheid, Michael and Rao, Rajesh PN and Brunton, Bingni W},
  journal={Scientific data},
  volume={9},
  number={1},
  pages={184},
  year={2022},
  publisher={Nature Publishing Group UK London}
}

@inproceedings{cui2024neurogpt,
  title={Neuro-gpt: Towards a foundation model for eeg},
  author={Cui, Wenhui and Jeong, Woojae and Th{\"o}lke, Philipp and Medani, Takfarinas and Jerbi, Karim and Joshi, Anand A and Leahy, Richard M},
  booktitle={2024 IEEE International Symposium on Biomedical Imaging (ISBI)},
  pages={1--5},
  year={2024},
  organization={IEEE}
}

@inproceedings{rasley2020deepspeed,
  title={Deepspeed: System optimizations enable training deep learning models with over 100 billion parameters},
  author={Rasley, Jeff and Rajbhandari, Samyam and Ruwase, Olatunji and He, Yuxiong},
  booktitle={Proceedings of the 26th ACM SIGKDD international conference on knowledge discovery \& data mining},
  pages={3505--3506},
  year={2020}
}

@article{mcinnes2018umap,
  title={Umap: Uniform manifold approximation and projection for dimension reduction},
  author={McInnes, Leland and Healy, John and Melville, James},
  journal={arXiv preprint arXiv:1802.03426},
  year={2018}
}

@article{mesgarani2014phonetic,
  title={Phonetic feature encoding in human superior temporal gyrus},
  author={Mesgarani, Nima and Cheung, Connie and Johnson, Keith and Chang, Edward F},
  journal={Science},
  year={2014}
}

@article{zahorodnii2025neuroprobe,
  title={Neuroprobe: Evaluating Intracranial Brain Responses to Naturalistic Stimuli},
  author={Zahorodnii, Andrii and Wang, Christopher and Stankovits, Bennett and Moraitaki, Charikleia and Chau, Geeling and Barbu, Andrei and Katz, Boris and Fiete, Ila R},
  journal={arXiv preprint arXiv:2509.21671},
  year={2025}
}

@article{abnar2020quantifying,
  title={Quantifying attention flow in transformers},
  author={Abnar, Samira and Zuidema, Willem},
  journal={arXiv preprint arXiv:2005.00928},
  year={2020}
}

@article{hickok2007cortical,
  title={The cortical organization of speech processing},
  author={Hickok, Gregory and Poeppel, David},
  journal={Nature reviews neuroscience},
  volume={8},
  number={5},
  pages={393--402},
  year={2007},
  publisher={Nature Publishing Group UK London}
}

@article{rauschecker2009maps,
  title={Maps and streams in the auditory cortex: nonhuman primates illuminate human speech processing},
  author={Rauschecker, Josef P and Scott, Sophie K},
  journal={Nature neuroscience},
  volume={12},
  number={6},
  pages={718--724},
  year={2009},
  publisher={Nature Publishing Group US New York}
}

@article{friederici2011brain,
  title={The brain basis of language processing: from structure to function},
  author={Friederici, Angela D},
  journal={Physiological reviews},
  volume={91},
  number={4},
  pages={1357--1392},
  year={2011},
  publisher={American Physiological Society Bethesda, MD}
}

@article{dehaene2011unique,
  title={The unique role of the visual word form area in reading},
  author={Dehaene, Stanislas and Cohen, Laurent},
  journal={Trends in cognitive sciences},
  volume={15},
  number={6},
  pages={254--262},
  year={2011},
  publisher={Elsevier}
}

@article{optuna,
  author       = {Takuya Akiba and
                  Shotaro Sano and
                  Toshihiko Yanase and
                  Takeru Ohta and
                  Masanori Koyama},
  title        = {Optuna: {A} Next-generation Hyperparameter Optimization Framework},
  journal      = {CoRR},
  volume       = {abs/1907.10902},
  year         = {2019},
  url          = {http://arxiv.org/abs/1907.10902},
  eprinttype    = {arXiv},
  eprint       = {1907.10902},
  bibsource    = {dblp computer science bibliography, https://dblp.org}
}

@article{Sharma2022Scaling,
  author  = {Utkarsh Sharma and Jared Kaplan},
  title   = {Scaling Laws from the Data Manifold Dimension},
  journal = {Journal of Machine Learning Research},
  year    = {2022},
  volume  = {23},
  number  = {9},
  pages   = {1--34},
  url     = {http://jmlr.org/papers/v23/20-1111.html}
}

@article{carzaniga2025swec,
  title={A foundation model with multi-variate parallel attention to generate neuronal activity},
  author={Carzaniga, Francesco and Hersche, Michael and Sebastian, Abu and Schindler, Kaspar and Rahimi, Abbas},
  journal={arXiv preprint arXiv:2506.20354},
  year={2025}
}

@inproceedings{
anonymous2026are,
  title={Are EEG foundation models worth it? comparative evaluation with traditional decoders in diverse BCI tasks},
  author={Yang, Liuyin and Sun, Qiang and Li, Ang and Van Hulle, Marc M},
  booktitle={The Fourteenth International Conference on Learning Representations},
  year={2026}
}

@article{oganesian2025barista,
  title={BaRISTA: Brain Scale Informed Spatiotemporal Representation of Human Intracranial Neural Activity},
  author={Oganesian, Lucine L and Hashemi, Saba and Shanechi, Maryam M},
  journal={arXiv preprint arXiv:2512.12135},
  year={2025}
}

@book{luck2014introduction,
  title={An Introduction to the Event-Related Potential Technique},
  author={Luck, Steven J.},
  edition={2},
  year={2014},
  publisher={MIT Press},
  url={https://mitpress.mit.edu/9780262525855/an-introduction-to-the-event-related-potential-technique/}
}

@book{cohen2014analyzing,
  title={Analyzing Neural Time Series Data: Theory and Practice},
  author={Cohen, Mike X.},
  year={2014},
  publisher={MIT Press},
  url={https://direct.mit.edu/books/monograph/4013/analyzing-neural-time-series-datatheory-and}
}

@article{buzsaki2012origin,
  title={The origin of extracellular fields and currents—EEG, ECoG, LFP and spikes},
  author={Buzs{\'a}ki, Gy{\"o}rgy and Anastassiou, Costas A and Koch, Christof},
  journal={Nature Reviews Neuroscience},
  volume={13},
  number={6},
  pages={407--420},
  year={2012},
  doi={10.1038/nrn3241}
}

@article{zheng2024discrete,
  title={Du-IN: Discrete units-guided mask modeling for decoding speech from intracranial neural signals},
  author={Zheng, Hui and Wang, Hai-Teng and Jiang, Wei-Bang and Chen, Zhong-Tao and He, Li and Lin, Pei-Yang and Wei, Peng-Hu and Zhao, Guo-Guang and Liu, Yun-Zhe},
  journal={Advances in Neural Information Processing Systems},
  volume={37},
  pages={79996--80033},
  year={2024}
}

@inproceedings{zheng-etal-2025-longhelpsshort,
  title = "When Long Helps Short: How Context Length in Supervised Fine-tuning Affects Behavior of Large Language Models",
  author = "Zheng, Yingming and
    Li, Hanqi and
    Yu, Kai and
    Chen, Lu",
  editor = "Christodoulopoulos, Christos and
    Chakraborty, Tanmoy and
    Rose, Carolyn and
    Peng, Violet",
  booktitle = "Proceedings of the 2025 Conference on Empirical Methods in Natural Language Processing",
  month = nov,
  year = "2025",
  address = "Suzhou, China",
  publisher = "Association for Computational Linguistics",
  url = "https://aclanthology.org/2025.emnlp-main.522/",
  doi = "10.18653/v1/2025.emnlp-main.522",
  pages = "10293--10308",
  isbn = "979-8-89176-332-6"
}

@inproceedings{gao-etal-2025-trainlongeffective,
  title = "How to Train Long-Context Language Models (Effectively)",
  author = "Gao, Tianyu and
    Wettig, Alexander and
    Yen, Howard and
    Chen, Danqi",
  editor = "Che, Wanxiang and
    Nabende, Joyce and
    Shutova, Ekaterina and
    Pilehvar, Mohammad Taher",
  booktitle = "Proceedings of the 63rd Annual Meeting of the Association for Computational Linguistics (Volume 1: Long Papers)",
  month = jul,
  year = "2025",
  address = "Vienna, Austria",
  publisher = "Association for Computational Linguistics",
  url = "https://aclanthology.org/2025.acl-long.366/",
  doi = "10.18653/v1/2025.acl-long.366",
  pages = "7376--7399",
  isbn = "979-8-89176-251-0"
}

@misc{shen2026brain4fms,
  title         = {{Brain4FMs}: A Benchmark of Foundation Models for Electrical Brain Signal},
  author        = {Shen, Fanqi and Yang, Enhong and Li, Jiahe and Hong, Junru and Pan, Xiaoran and Yuan, Zhizhang and Li, Meng and Yang, Yang},
  year          = {2026},
  eprint        = {2602.11558},
  archivePrefix = {arXiv},
  primaryClass  = {cs.LG},
  doi           = {10.48550/arXiv.2602.11558},
  url           = {https://arxiv.org/abs/2602.11558}
}

@misc{liu2026eegfoundationmodels,
  title         = {{EEG} Foundation Models: Progresses, Benchmarking, and Open Problems},
  author        = {Liu, Dingkun and Chen, Yuheng and Chen, Zhu and Cui, Zhenyao and Wen, Yaozhi and An, Jiayu and Luo, Jingwei and Wu, Dongrui},
  year          = {2026},
  eprint        = {2601.17883},
  archivePrefix = {arXiv},
  primaryClass  = {cs.LG},
  doi           = {10.48550/arXiv.2601.17883},
  url           = {https://arxiv.org/abs/2601.17883}
}

@misc{banville2025scalinglaws,
  title         = {Scaling Laws for Decoding Images from Brain Activity},
  author        = {Banville, Hubert and Benchetrit, Yohann and d'Ascoli, St{\'e}phane and Rapin, J{\'e}r{\'e}my and King, Jean-R{\'e}mi},
  year          = {2025},
  eprint        = {2501.15322},
  archivePrefix = {arXiv},
  primaryClass  = {eess.IV},
  doi           = {10.48550/arXiv.2501.15322},
  url           = {https://arxiv.org/abs/2501.15322}
}

@misc{bomatter2025limitedparticipantdiversity,
  title         = {Is Limited Participant Diversity Impeding {EEG}-based Machine Learning?},
  author        = {Bomatter, Philipp and Gouk, Henry},
  year          = {2025},
  eprint        = {2503.13497},
  archivePrefix = {arXiv},
  primaryClass  = {eess.SP},
  doi           = {10.48550/arXiv.2503.13497},
  url           = {https://arxiv.org/abs/2503.13497}
}

@article{card_speech_2024,
  author  = {Card, Nicholas S. and Wairagkar, Maitreyee and others and Stavisky, Sergey D. and Brandman, David M.},
  title   = {An accurate and rapidly calibrating speech neuroprosthesis},
  journal = {New England Journal of Medicine},
  volume  = {391},
  number  = {7},
  pages   = {609--618},
  year    = {2024},
  doi     = {10.1056/NEJMoa2314132}
}

@article{wairagkar_voice_2025,
  author  = {Wairagkar, Maitreyee and Card, Nicholas S. and Singer-Clark, Tyler and others and Brandman, David M. and Stavisky, Sergey D.},
  title   = {An instantaneous voice-synthesis neuroprosthesis},
  journal = {Nature},
  volume  = {644},
  number  = {8075},
  pages   = {145--152},
  year    = {2025},
  doi     = {10.1038/s41586-025-09127-3}
}

\end{document}